\documentclass[lettersize,journal]{IEEEtran}
\usepackage{amsmath,amsfonts}
\usepackage{algorithmic}
\usepackage{algorithm}
\usepackage{array}
\usepackage[caption=false,font=normalsize,labelfont=sf,textfont=sf]{subfig}
\usepackage{textcomp}
\usepackage{stfloats}
\usepackage{url}
\usepackage{verbatim}
\usepackage{graphicx}
\usepackage{cite}
\usepackage{multirow}
\usepackage{booktabs}
\usepackage{color}
\usepackage{orcidlink}\hypersetup{colorlinks=true,linkcolor=black}

\begin{document}

\title{Learn Temporal Consistency For Robust Satellite Video Detector}
\author{Weilong Guo~\orcidlink{0000-0002-9259-1746}, Shengyang Li~\orcidlink{0000-0002-9888-9869}, and Yanfeng Gu, \IEEEmembership{Senior Member,~IEEE,}
\thanks{\textit{Corresponding author: Shengyang Li.}.}
\thanks{Weilong Guo and Shengyang Li is with the Technology and Engineering Center for Space Utilization, Chinese Academy of Sciences, Beijing 100094, China 
and Key Laboratory of Space Utilization, Chinese Academy of Sciences, Beijing 100094, China. 
Shengyang Li is also with the University of Chinese Academy of Sciences, Beijing 100049, China 
(e-mail: guoweilong19@mails.ucas.ac.cn; shyli@csu.ac.cn).}
\thanks{Yanfeng Gu is with the School of Electronics and Information Engineering, Harbin Institute of Technology, Harbin 150001, 
China (e-mail: guyf@hit.edu.cn).}
}

\markboth{Journal of \LaTeX\ Class Files,~Vol.~14, No.~8, August~2021}%
{Shell \MakeLowercase{\textit{et al.}}: A Sample Article Using IEEEtran.cls for IEEE Journals}


\maketitle

\begin{abstract}
Satellite video object detection (SVOD) for oriented and fine-grained objects plays an important role in satellite applications. Most existing SVOD methods only focus on one or a few coarse-grained categories of moving objects and represent objects with horizontal bounding boxes. They have difficulty extracting complete, accurate, and consistent information about objects in whole satellite videos. In this paper, we propose a satellite video object detection framework based on Temporal Consistency Learning (TCL). TCL adeptly detects oriented and fine-grained objects by leveraging the rich temporal contexts within satellite videos. The framework integrates three key modules: temporal and fine-grained feature aggregation (TFA), structure encoding (SE), and temporal consistency constraint (TCC). TFA and TCC modules facilitate consistent representation learning across frames, while the SE module encodes both appearance and structural information for precise fine-grained recognition. Experimental results on the SAT-MTB benchmark dataset demonstrate TCL's superior performance, achieving a new state-of-the-art oriented and fine-grained detection accuracy of 47.7$\%$ mAP—a 4.8$\%$ improvement over the baseline. Furthermore, our TCL framework readily accommodates existing image-based detectors, leading to enhanced detection accuracies.
\end{abstract}


\section{Introduction}

\IEEEPARstart{V}{ideo} object detection, centered on the identification and localization of objects in dynamic sequences~\cite{zhou2023transvod}, holds a pivotal role in various fields, including intelligent video surveillance~\cite{jiao2021new}, security equipment~\cite{zheng2021survey}, and video comprehension~\cite{zhao2019obj}. The advent of video satellites has empowered the acquisition of continuous observational videos, offering a wealth of spatio-temporal data for specific ground regions~\cite{zhao2022satsot}. The expansive scope and rich information in satellite videos render them invaluable for applications like disaster response, ocean observation, and smart cities~\cite{li2023mtb,gu2020svsc}. In these scenarios, precise object localization within satellite videos, referred to as the Satellite Video Object Detection (SVOD) task, becomes crucial.

The intersection of video processing and deep learning has catalyzed the evolution of SVOD~\cite{FENG2021116,s23125771,Feng2023SDANetSD}.
Previous research has predominantly focused on Moving Object Detection (MOD) within satellite videos, with a particular emphasis on moving entities such as vehicles~\cite{li2023satreview,Xiao2023IncorporatingDB,li2023mtb}.
Conventional MOD methods typically frame the task as foreground and background separation, employing techniques like background modeling~\cite{Zhang2019ErrorBF,Zhang2019OnlineSS,Zhang2021MovingVD,Lei2021TinyMV} and frame difference~\cite{Zhang2017SpaceOD,Li2019ShipDA,Shi2020DetectingAT,Shu2021SmallMV,Chen2020ANA} to capture moving foreground elements. While recent contributions in deep learning methods have effectively harnessed temporal contexts in satellite videos for MOD~\cite{FENG2021116,Liu2018LowQualityAM,Xiao2022DSFNetDA,Zhou2022FewShotAD,Pi2022VeryLM}, the exclusive focus on moving objects presents practical challenges:
1) Objects are not perpetually in motion, and MOD methods often struggle to model the complete patterns of objects in satellite videos.
2) MOD approaches tend to specialize in a single category of objects, neglecting the fine-grained details present in satellite videos.
3) Current MOD methods typically represent objects using Horizontal Bounding Boxes (HBBs), 
despite the fact that these objects are inherently arbitrary-oriented.

\begin{figure}[!t]
  \centering
  \includegraphics[width=3.5in]{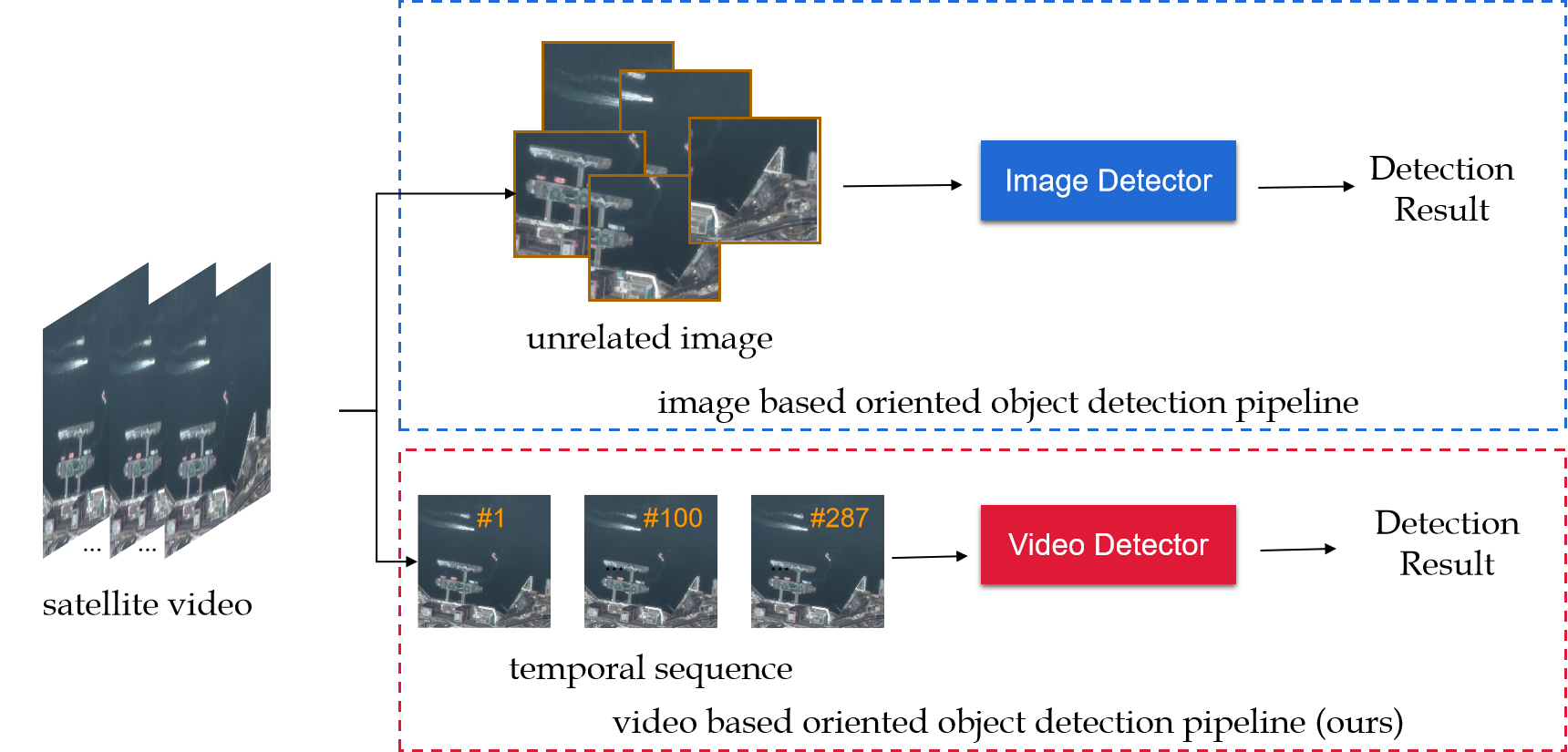}
  \caption{Image detectors vs Video detectors for oriented objects of satellite videos. 
  (1) The image detectors treat satellite videos as a series of unrelated images and destroy their temporal structures.
  (2) The video detectors can utilize the rich temporal contexts of satellite videos, helping improve detection performance.}
  \label{figivsv}
\end{figure}

The release of SAT-MTB, the inaugural large-scale and open satellite video benchmark dataset~\cite{li2023mtb}, has underscored the need for innovative video detectors capable of handling oriented and fine-grained objects. Despite this imperative, current knowledge indicates the absence of any video detector capable of simultaneously addressing the challenges posed by fine-grained and arbitrarily oriented objects in satellite videos~\cite{li2023satreview}.

Despite the strides made by image-based detectors for oriented objects, leveraging datasets like DOTA~\cite{Ding2021ObjectDI}, DIOR-R~\cite{Cheng2021AnchorFreeOP}, and FAIR1M~\cite{Sun2021FAIR1MAB}, 
these detectors face challenges when applied to satellite videos. 
As shown in Figure~\ref{figivsv}, although capable of recognizing oriented objects in the context of individual frames, these detectors treat satellite videos as disjointed images, disregarding their temporal structure. This approach leads to fragmented object descriptions, introducing inconsistencies in timing. Consequently, image-based detectors struggle to achieve optimal performance on satellite videos~\cite{li2023satreview,li2023mtb}. There arises a need for a video detector designed for both oriented and fine-grained objects, one that can effectively mine the rich temporal contexts inherent in satellite videos.

To tackle this challenge, we propose a novel satellite video object detection framework based on Temporal Consistency Learning (TCL). 
This framework is designed to detect both oriented and fine-grained objects and comprises three key modules: temporal and fine-grained feature aggregation (TFA), structure encoding (SE), and temporal consistency constraint (TCC). TFA enhances the fine-grained features of objects by leveraging information from adjacent frames, fostering temporal consistency in the model's understanding of the same object. 
The SE module encodes both appearance and structural information, 
providing a comprehensive representation for accurate fine-grained recognition. Meanwhile, TCC formulates an objective function that compels features of the same object across consecutive frames to remain consistent. Notably, TCL stands out as the pioneering video object detection framework tailored for oriented and fine-grained objects in satellite videos~\cite{li2023mtb,li2023satreview}. This framework not only supports the mining of temporal contexts but also captures the complete patterns of moving and static objects throughout the entire satellite video sequence. Unlike image-based detectors that treat videos as disjointed frames, TCL offers a holistic perspective on oriented and fine-grained objects, ushering in a new paradigm in satellite video object detection.

The primary contributions of this paper are summarized as follows:

\begin{enumerate}
  \item {Designing a Novel Framework: We introduce the first-ever satellite video object detection framework tailored for 
  oriented and fine-grained objects, capable of mining temporal contexts.}
  \item {Temporal Consistency Learning Strategies: Our proposed Temporal and Fine-Grained Feature Aggregation (TFA) and 
  Temporal Consistency Constraint (TCC) strategies contribute to capturing complete and consistent patterns of 
  oriented and fine-grained objects throughout entire satellite videos.}
  \item {Structure Information Using: The Structure Encoding (SE) module is devised to encode both appearance and 
  structural information of objects, enhancing the precision of fine-grained recognition.}
  \item {Benchmark Leadership: Through extensive comparison experiments and comprehensive ablation studies on 
  the largest satellite video benchmark dataset, our TCL framework achieves top-tier performance, refreshing existing leaderboards.}
\end{enumerate}

\section{Related Works}
In this section, we mainly review the satellite video moving object detection based on deep learning method, oriented object detection of satellite images, and video object detection method of general videos.

\subsection{Satellite Video Moving Object Detection based on Deep Learning}
Most satellite video moving object detection methods 
based on deep learning~\cite{FENG2021116,Liu2018LowQualityAM,Xiao2022DSFNetDA,Zhou2022FewShotAD,Pi2022VeryLM} are dedicated to mining the temporal contexts of satellite videos, also called motion information in some works, 
to compensate for the moving small objects with weak appearance~\cite{li2023satreview}.

ClusterNet~\cite{LaLonde2017ClusterNetDS} proposed a novel two-stage spatio-temporal CNN. 
In the first stage, ClusterNet considers both the motion and appearance information and 
combines them within the convolution architecture. 
It also develops the regions of objects of interest (ROOBI). 
In the second stage, it estimates the center location of objects in the ROOBI. 
The detector of~\cite{Li2019WeakMO} computes the difference map between two adjacent frames as temporal contexts 
and stacks it with the original RGB channel 
so that a multiple-channel input is obtained. 
Then, a motion-driven detection network is constructed to obtain the fusion features 
from the multiple-channel input. 
CKDNet~\cite{FENG2021116} proposes a cross-frame keypoint-based detection network. 
It develops the cross-frame module and extracts apparent-temporal fusion features by cross-difference operation. 
Chi et al.~\cite{Chi2020AerialVM} uses ShuffleNet unit~\cite{Zhang2017ShuffleNetAE} and 
long-short-term memory (LSTM) network to capture temporal contexts. 
As a result, the neighboring frames can be used to improve the detection accuracy of the current frame. 
Pi et al.~\cite{Pi2022VeryLM} design a differential module in the stage of feature extraction. 
It integrates the motion information from adjacent frames to facilitate the extraction of effective semantic features. 
DTSTC~\cite{s23125771} enhances the precision by fusing road masks from the spatial domain with motion heat maps from the temporal domain. 
SDANet~\cite{Feng2023SDANetSD} segments the road in vast scenes and embeds its semantic features in the network by weakly supervised learning, 
which guides the detector to emphasize the regions of interest. 
It reduces the false detection caused by massive interference.

Most of these methods only focus on small moving cars~\cite{li2023satreview}. They cannot capture the complete patterns of diverse objects in satellite videos. In this paper, our TCL is designed to fill these gaps.

\subsection{Satellite image oriented object detection}
In satellite images, also called remote sensing images, objects are with arbitrary orientations. A horizontal bounding box (HBB) cannot depict the object orientation and contains redundant information of the background~\cite{Wang2023OrientedOD}. To cope with these challenges, more efforts have been devoted to oriented object detection. They represent these objects with oriented bounding boxes (OBBs). Existing methods can be classified into three categories: 
detection based on oriented object representation 
(DB-OOR)~\cite{Xu2019GlidingVO,Wang2020LearningCP,yang2021rethinking,yang2021learning,yang2022kfiou,Li2021OrientedRF,Guo2021BeyondBC,Hou2022GRepGR}, detection based on feature alignment (DB-FA)~\cite{Ding2019LearningRT,Xie2021OrientedRF,Yang2019R3DetRS,Han2020AlignDF,Pan2020DynamicRN}, and detection based on rotation-invariant feature learning (DB-RIFL)~\cite{Han2021ReDetAR}.

The DB-OOR methods are dedicated to exploring mathematical models suitable for oriented objects. Gliding vertex~\cite{Xu2019GlidingVO} and CenterMap~\cite{Wang2020LearningCP} use quadrilateral and mask to accurately describe oriented objects, respectively.  GWD~\cite{yang2021rethinking}, KLD~\cite{yang2021learning}, and KFIoU~\cite{yang2022kfiou} represent oriented objects as a 2-D Gaussian distribution. GWD and KLD calculate the the Gaussian Wasserstein distance and Kullback-Leibler Divergence, as loss separately. KFIoU proposes an effective approximate SkeIoU loss on Gaussian modeling and Kalman filter. Oriented RepPoints~\cite{Li2021OrientedRF} represent objects as a set of sample points. CFA~\cite{Guo2021BeyondBC} models irregular object layout and shape as convex hulls. G-Rep~\cite{Hou2022GRepGR} proposes a unified Gaussian representation to construct Gaussian distributions for oriented bounding boxes, quadrilateral bounding boxes, and point sets. Some methods~\cite{Azimi2018TowardsMO,Ma2017ArbitraryOrientedST,Zhang2018TowardAS} also adopt rotated anchors with different angles, scales, and aspect ratios for better representation of oriented objects.

The DB-FA methods align features between receptive fields and oriented objects. 
RoI Transformer~\cite{Ding2019LearningRT} transforms Horizontal RoIs (HRoIs) into RRoIs, 
facilitating the extraction of accurate features from oriented objects.
Oriented RCNN~\cite{Xie2021OrientedRF} directly learns oriented proposals for extracting features from oriented regions. 
R3Det~\cite{Yang2019R3DetRS} and S2A-Net~\cite{Han2020AlignDF} employ feature alignment techniques 
between horizontal receptive fields and rotated anchors.
DRN~\cite{Pan2020DynamicRN} utilizes dynamic feature selection and refinement processes to capture aligned features.

The DB-RIFL methods are designed to learn feature representations that remain invariant under rotations.
Regardless of the rotation transformations applied to the input, the output remains consistent. 
In the classic ReDet~\cite{Han2021ReDetAR}, improvements to the feature representation are propagated throughout the network. 
This results in rotation-equivariant features within the backbone network.
In its detection head, the proposed RiRoI Align extracts completely rotation-invariant features for robust object localization.

Additionally, several methods address other specific challenges, 
such as the discontinuous boundaries problem~\cite{yang2020arbitrary} and 
the large aspect ratio problem~\cite{Hou2022ShapeAdaptiveSA}. 
Furthermore, there are end-to-end transformer methods~\cite{Dai2022AO2DETRAO} that have demonstrated success 
in overcoming these challenges. 
These approaches have achieved impressive detection results across various open satellite image datasets.

While treating satellite videos as a series of unrelated images allows for their direct application, 
these image-based approaches overlooks the rich temporal contexts inherent in satellite videos. 
Consequently, these methods struggle to achieve optimal detection performance on satellite videos. 
Enter our TCL, a video object detection framework designed specifically for oriented and fine-grained objects. 
TCL excels by simultaneously supporting the mining of temporal contexts, 
ensuring enhanced performance in the detection of objects within satellite videos.

\subsection{Video Object Detection For General Videos}
Existing general video object detection methods primarily address temporal context in two distinct manners: 
post-processing~\cite{Han2016SeqNMSFV,Kang2016TCNNTW,Belhassen2019ImprovingVO,Sabater2020RobustAE} and 
feature aggregation of temporal 
information~\cite{Yao2020VideoOD,Jiang2019LearningWT,Han2020MiningIP,Han2020ExploitingBF,Lin2020DualSF,He2020TemporalCE,Chen2018OptimizingVO}.

Post-processing-based methods typically employ an image detector to acquire detection results for each frame. 
Subsequently, these results are associated through various post-processing strategies to enhance their coherence and stability. 
Notable examples include Seq-NMS~\cite{Han2016SeqNMSFV} and T-CNN~\cite{Kang2016TCNNTW}.

Feature aggregation-based methods enhance per-frame features by aggregating features from nearby frames. 
These methods often leverage additional components, 
such as optical flow models~\cite{Wang2018FullyMN,Zhu2017TowardsHP,jin2022feature}, 
recurrent neural networks~\cite{Deng2019ObjectGE,Guo2019ProgressiveSL}, 
deformable convolution fusion~\cite{He2020TemporalCE,Jiang2019VideoOD}, 
and relation networks~\cite{Chen2020MemoryEG,Deng2019RelationDN}. 
Another line of approaches~\cite{Chen2020MemoryEG,Wu2019SequenceLS} 
utilizes self-attention~\cite{Vaswani2017AttentionIA} and non-local methods~\cite{Wang2017NonlocalNN} 
to capture long-range dependencies within temporal contexts.
Several works~\cite{Yao2020VideoOD,Jiang2019LearningWT,Li2021ImprovingVI} 
focus on real-time video object detection while maintaining or improving accuracy. 
A recent method, TransVOD~\cite{He2022TransVODEV}, designs the video object detection architecture 
based on a spatial-temporal transformer architecture, achieving outstanding detection performance.

While all these methods are tailored for general videos and typically represent objects with horizontal bounding boxes, 
such an approach may not be optimal for satellite videos. 
In this paper, we introduce a novel TCL method, featuring a video object detection architecture 
specifically designed to detect oriented and fine-grained objects in satellite videos.


\subsection{Overall Framework}
The overall framework of our TCL is illustrated in figure~\ref{figmtd}. 
Given an input satellite video $\mathcal{V}$, our framework operates as follows:
(1) Each frame undergoes the feature extraction module, resulting in the extraction of its feature $F$.
(2) The oriented region proposal network (ORPN) predicts and extracts features of oriented regions of interest (ORoIs), 
denoted as $\Omega ^t$, based on the extracted feature $F$.
(3) The temporal and fine-grained feature aggregation (TFA) module enhances fine-grained features of objects in the current frame $f_t$ 
by leveraging those of objects in nearby frames. This process yields enhanced features of objects, denoted as $\Omega ^{t+}$.
(4) The enhanced features are then fed into the prediction head with structure encoding (SE) 
to predict the locations and categories of objects (4-a). 
Additionally, a temporal consistency constraint (TCC) loss function is designed to 
ensure that features of the same object remain consistent across continuous frames (4-b).

\subsection{Feature Extraction}
We employ the ReResNet50 + ReFPN architecture~\cite{Han2021ReDetAR} as our feature extraction module to extract features 
from each frame of the satellite video, $\mathcal{V}$, for oriented and fine-grained object detection. 
While existing video detectors designed for general videos assume that objects have consistent scales 
and typically use features from the last layer of the backbone for detection~\cite{jiao2021new,Han2020MiningIP}, 
this assumption is not applicable to satellite videos, where significant scale variations exist~\cite{li2023satreview}.
In this paper, we leverage five features outputted by the feature extraction module to detect objects of different scales. 
These features have 256 channels, and their receptive fields are \{4, 8, 16, 32, 64\}, respectively.

\subsection{Oriented Region Proposal Network}
\begin{figure}[!htbp]
  \centering
  \includegraphics[width=3.3in]{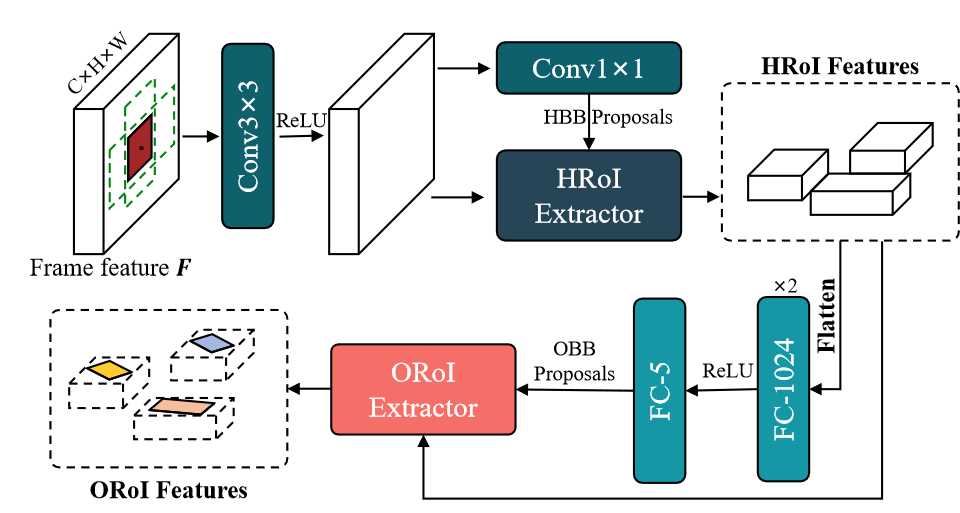}
  \caption{Pipeline of Oriented Region Proposal Network (ORPN).}
  \label{figorpn}
\end{figure}

Recent advancements in video object detection have demonstrated the effectiveness of mining temporal contexts 
to improve detection performance~\cite{Han2020MiningIP,zhou2023transvod}. 
However, existing methods often extract features from horizontal region proposals, 
assuming objects with a consistent orientation~\cite{jiao2021new,li2023satreview}. 
This approach proves unsuitable for oriented objects in satellite videos, 
resulting in features that include noisy background information. 
Inspired by RoI Transformer~\cite{Ding2019LearningRT}, we introduce the oriented region proposal network (ORPN) 
to extract more accurate features of oriented objects in satellite videos.  

As depicted in figure~\ref{figorpn}, the frame feature $F$ is input into the ORPN to predict and extract features of oriented objects. 
The process involves two stages: (1) In the first stage, a $3\times 3$ convolution with ReLU activation function 
and a $1\times 1$ convolution predict horizontal proposals for oriented objects. 
The HRoI (Horizontal Region of Interest) Extractor extracts their features as HRoIs based on these proposal locations.
(2) In the second stage, the HRoI features are flattened to vectors, and two fully connected layers with ReLU activation 
are employed to model dependencies between local features and predict proposals represented by oriented bounding boxes. 
The ORoI (Oriented Region of Interest) Extractor extracts accurate features of oriented objects, 
denoted as $\Omega ^t$, based on these oriented proposals.
It's important to note that figure~\ref{figorpn} illustrates only the regression branch in each stage. 
In practice, each regression branch corresponds to a binary classification branch for distinguishing between the foreground and background. 
The detailed implementation of HRoI Extractor and ORoI Extractor aligns with RoI Transformer~\cite{Ding2019LearningRT}.


Satellite videos inherently contain an abundance of redundant temporal information, 
a valuable prior widely utilized in satellite video interpretation tasks, including object detection, tracking, 
and segmentation~\cite{li2023satreview,li2023mtb}.
The same object usually appears in adjacent frames. 
Leveraging this knowledge, we design the Temporal and Fine-grained Feature Aggregation module (TFA) 
to enhance the fine-grained features of oriented objects.

The inputs of our TFA are the ORoI features of current frame $f_t$, denoted as $\Omega^t$, 
and neighbor frames $\{{f_{t-l},\ldots ,f_{t+l}}$\}, denoted as $\Omega^{ref}$.
We partition the ORoI feature of each object into $N_b$ blocks, each representing a local and fine-grained semantic aspect of the object, 
where $\Omega^t=\{\Omega^t_1,\ldots,\Omega^t_b,\ldots,\Omega^t_{N_b}\}$ 
and $\Omega^{ref}=\{\Omega^{ref}_1,\ldots,\Omega^{ref}b,\ldots,\Omega^{ref}{N_b}\}$. 
For instance, as shown in figure~\ref{figtfas}, when dividing an aircraft into four blocks—left wing, body, right wing, and tail—each block 
encapsulates a specific aspect of the object.
Here, we illustrate the TFA pipeline with the aggregation of b-th local and fine-grained features. 
For b-th fine-grained feature of any object of current frame, $C_u\in \Omega^t_b$, and that of neighbor frames, $C_v\in \Omega^{ref}_b$, 
we model their relations as follows:

\begin{equation}
  \label{eq1}
  \mathcal{R} _u = \frac{e^{[(w_u*C_u)\odot (w_v*C_v)]}}{\sum_{1\leq v\leq N_{ref}}e^{[(w_u*C_u)\odot (w_v*C_v)]} }
\end{equation}

Here, $\odot $ represents  the inner product of feature vectors. $N_{ref}$ is the total number of objects in neighboring frames.
$w_u$ and $w_v$ are learnable weights from fully connected layers.
$*$ means element-wise multiplication. Then we aggregate fine-grained features for $C_u$ based on these relations, 
obtaining enhanced fine-grained feature $C_u^{+}$.
\begin{equation}
  \label{eq2}
  C_u^{+} = \sum_{1\leq v\leq N_{ref}} w_u^{+}*\mathcal{R} _u*C_v
\end{equation}

$w_u^{+}$ denotes the learnable weight of the enhanced fine-grained feature, implemented through a fully connected layer.
Repeating the above pipeline for each local and fine-grained feature of each object in current frame 
achieves temporal fine-grained feature aggregation.
Stacking the enhanced fine-grained features together yields the fine-grained feature-enhanced object feature $R_i^{t+}\in \Omega^{t+}$.

\subsection{Structure Encoding}
\begin{figure}[!htbp]
  \centering
  \includegraphics[width=3.3in]{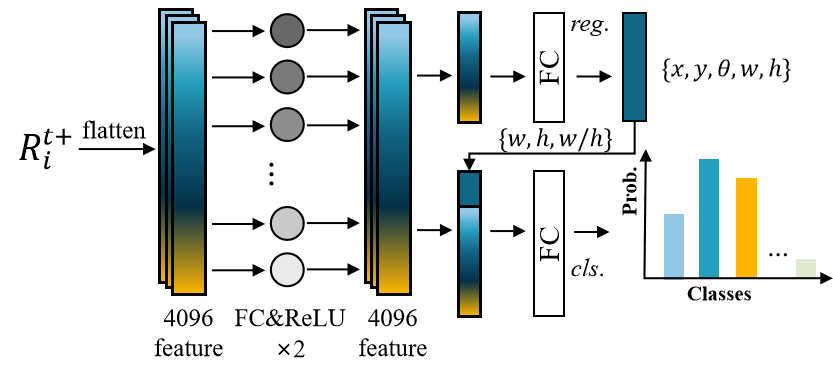}
  \caption{Pipeline of prediction head with structure encoding (SE).}
  \label{figse}
\end{figure}

Deriving proper and fine-grained representations from limited and distorted appearance information of objects in satellite videos 
poses a significant challenge. However, structural information about objects, including width, length, and aspect ratio, 
often serves as potent discriminative information. Motivated by this insight, 
we introduce the Prediction Head with Structure Encoding (SE), 
a module that leverages both appearance and structural information to achieve accurate and fine-grained detection. 
Detailed insights into the SE module are provided in figure~\ref{figse}.

The input to the SE module is the enhanced object feature $R_i^{t+}$. 
After flattening, we obtain a $1\times 4096$ feature vector. This vector is then fed into two fully connected (FC) layers 
with ReLU activation functions, producing a feature vector, $F_p$, used for both regression and fine-grained classification.
For the regression branch, another FC layer is employed to predict the object's location information, 
denoted as $\{x, y, \theta, w, h\}$. Here, $x$ and $y$ represent the position of the object's center point, 
$\theta$ signifies the rotation angle, and $w$ and $h$ denote the width and height of the object. 
Additionally, we stack the structural information $\{w, h, w/h\}$ with the feature $F_p$. 
This combined information is input into another FC layer to predict the fine-grained category of the object.

\subsection{Temporal Consistency Constraint}
The objective function of the Temporal Consistency Constraint (TCC) loss aims to 
enforce the consistency of features for the same object across consecutive frames. 
The inputs to the TCC loss are the object features of the current frame, $\Omega^t$, 
and those of neighboring frames, $\Omega^{ref}$. 
For any $R_i^t\in \Omega^t$ and $R_j^{ref}\in \Omega^{ref}$, we employ the Euclidean distance to quantify the similarity between them:
\begin{equation}
  \label{eq3}
  dist_{ij} = \|\|R_i^t-R_j^{ref} \|  \| _2^2 = \sum _{k=1}^C(R_{ik}^t-R_{jk}^{ref})^2
\end{equation}

The variable $dist$ represents the measured $N_t \times N_{ref}$ distance matrix between objects in $\Omega^t$ and $\Omega^{ref}$. 
Here, $N_t$ and $N_{ref}$ denote the number of objects in $\Omega^t$ and $\Omega^{ref}$, respectively.
To construct the ground truth of the distance matrix, $G_{sim}$, for calculating the loss, 
we utilize the fine-grained labels of objects. 
Specifically, the distance between objects with the same fine-grained labels is set to 1, 
while the distance between objects with different fine-grained labels is set to 0. 
The loss can be represented as:
\begin{equation}
  \label{eq4}
  \mathcal{L} _{fine} = G_{sim}*(dist-\delta _f^p)^2+(1-G_{sim})*max(\delta _f^q-dist,0)^2
\end{equation}

The terms $\delta _f^p$ and $\delta _f^q$ represent dynamic thresholds for the distances between objects of 
the same categories and different categories, respectively. 
These thresholds dynamically adapt to the feature distribution of objects across various fine-grained categories. 
In practical implementation,  
we use the average of the $dist$ matrix for the distances between objects of the same categories as $\delta _f^p$   
and that of different categories as $\delta _f^q$. 
In the context of satellite videos, objects with weak appearances often exhibit strong similarity with the background. 
To prevent the detection from overlooking these objects, 
we extend the aforementioned loss function to encompass both foreground and background features. 
In this extension, we categorize all objects of interest as foreground and treat others as background. 
The loss function can be represented as:
\begin{equation}
  \label{eq5}
  \mathcal{L} _{obj} = (1-G_{sim}^{obj})*max(\delta _{obj}^b-dist,0)^2
\end{equation}

We employ a consistent strategy to construct the ground truth of the distance matrix $G_{sim}^{obj}$ 
for both the foreground and the background. 
When the labels of two features are both the foreground or background, the distance is 1 and 0 otherwise. 
$\delta _{obj}^b$ stands for the dynamic threshold for the distances between the foreground and background. 
In each iteration, we use the maximum distance between the foreground and the background features as $\delta _{obj}^b$. 
The final TCC loss function can be represented as:
\begin{equation}
  \label{eq6}
  \mathcal{L} _{tcc} = \mathcal{L} _{fine} + \mathcal{L} _{obj}
\end{equation}

\subsection{Loss Function}
The total loss $\mathcal{L}$ of our TCL consists of the losses of ORPN, SE, and TCC.
\begin{equation}
  \label{eq7}
  \mathcal{L} = \mathcal{L} _{orpn} + \mathcal{L} _{se} + \lambda \mathcal{L} _{tcc}
\end{equation}

The losses for the Oriented Region Proposal Network (ORPN), 
Prediction Head with Structure Encoding (SE), and Temporal Consistency Constraint (TCC) 
are denoted as $\mathcal{L} _{orpn}$, $\mathcal{L} _{se}$, and $\mathcal{L} _{tcc}$, respectively. 
Both ORPN and SE encompass regression and fine-grained classification losses. 
The regression loss function for these components is the smooth L1 loss, 
while the classification loss function is the cross-entropy loss.

\section{Experiments}
To assess the effectiveness of our proposed method, 
we undertake experiments on the SAT-MTB, the largest satellite video multi-mission benchmark dataset~\cite{li2023mtb}. 
Initially, we perform a comparative analysis against state-of-the-art image-based oriented object detection methods specifically 
on the SAT-MTB-OBB sub-dataset. 
Subsequently, a series of ablation experiments are conducted on the dataset to systematically evaluate 
the efficacy of each module in our proposed method.
To provide a comprehensive comparison with state-of-the-art video detectors, 
we further extend our experiments to the SAT-MTB-HBB sub-dataset. 
This extension allows us to assess the superiority of our method in leveraging temporal contexts for improved performance.

\subsection{Comparison Methods}
We compare our TCL with methods that are already included in the SAT-MTB-OBB benchmark and other recent SOTA oriented object detection methods 
to demonstrate its effectiveness in oriented and fine-grained object detection. 
It includes Faster RCNN-OBB~\cite{Ren2015FasterRT}, RetinaNet-OBB~\cite{Lin2017FocalLF}, 
RoI Transformer~\cite{Ding2019LearningRT}, FCOS-OBB~\cite{Tian2019FCOSFC}, 
RepPoints-OBB~\cite{Yang2019RepPointsPS}, ATSS-OBB~\cite{Zhang2019BridgingTG}, Oriented RCNN~\cite{Xie2021OrientedRF}, 
R3Det~\cite{Yang2019R3DetRS}, S2ANet~\cite{Han2020AlignDF}, Gliding Vertex~\cite{Xu2019GlidingVO}, 
GWD~\cite{yang2021rethinking}, KFIoU~\cite{yang2022kfiou}, KLD~\cite{yang2021learning}, 
SASM~\cite{Hou2022ShapeAdaptiveSA}, and ReDet~\cite{Han2021ReDetAR}. 
The -OBB stands for the methods designed for HBB object detection. 
Their OBB versions here are re-implemented by the open project mmrotate\footnote{\label{mmrt}https://github.com/open-mmlab/mmrotate}.

To evaluate the superiority of our TCL in the use of temporal contexts of satellite videos, 
we compare it with 4 SOTA video object detectors 
(e.g. FGFA~\cite{zhu2017fgfa}, DFF~\cite{zhu2017dff}, SELSA~\cite{Wu2019SequenceLS}, TemporalRoI Align~\cite{gong2021selsa}) 
and 13 advanced image-based object detectors 
(e.g. Faster RCNN~\cite{Ren2015FasterRT}, YOLOv3~\cite{Redmon2018YOLOv3AI}, 
Grid RCNN~\cite{Lu2018GridR}, TridentNet~\cite{Li2019ScaleAwareTN}, 
FCOS~\cite{Tian2019FCOSFC}, Guided Anchoring~\cite{Wang2019RegionPB}, 
RetinaNet~\cite{Lin2017FocalLF}, DETR~\cite{Carion2020EndtoEndOD}, 
ATSS~\cite{Zhang2019BridgingTG}, Libra RCNN~\cite{Pang2019LibraRT}, 
Groie~\cite{Rossi2020ANR}, Cascade RCNN~\cite{Cai2019CascadeRH}, DINO~\cite{Zhang2022DINODW}). 
Their code are from mmtracking\footnote{\label{mmtk}https://github.com/open-mmlab/mmtracking}
 and mmdetection\footnote{\label{mmdt}https://github.com/open-mmlab/mmdetection} open project.

\subsection{Dataset and Evaluation Metrics}

SAT-MTB is the largest satellite video dataset published in 2023~\cite{li2023mtb}, which supports multiple tasks.
The detection tasks for fine-grained objects represented with oriented bounding boxes and horizontal bounding boxes 
are marked with OBB and HBB respectively.
Here, we denote the OBB and HBB fine-grained object detection sub-dataset of SAT-MTB as SAT-MTB-OBB and SAT-MTB-HBB, respectively.
The SAT-MTB-OBB contains 106 videos, 22767 frames, and 211425 instances. 
The SAT-MTB-HBB contains 144 videos, 33228 frames, and 308204 instances.
They are consisted of 12 fine-grained categories: wide-bodied aircraft (WA), narrow-bodied aircraft (NA), rear-engined aircraft (RA), 
four-engine aircraft (FA), corporate aircraft (CA), speed boat (SB), yacht (YH), cruise (CS), freighter (FH), 
naval vessel (NV), other ship (OS), and train (TN). 
Some samples of SAT-MTB-OBB with OBB labels are shown in figure~\ref{figds}. 
For the detection accuracy, we adopt the mean average precision (mAP, IoU=0.5) as the main evaluation metric, which is consistent with the SAT-MTB benchmark.
We also use the average precision (AP, IoU=0.5) to evaluate the detection performance of each fine-grained category.

\subsection{Implement Details}
We implement the proposed method TCL in PyTorch.
All codes of comparison methods are from mmrotate$^{~\ref{mmrt}}$, 
mmdetection$^{~\ref{mmdt}}$, 
and mmtracking$^{~\ref{mmtk}}$ open projects.
All experiments are conducted on a single NVIDIA Tesla V100 GPU with 32 GB memory.
It is noted that, in the official OBB benchmark of SAT-MTB, satellite videos are treated as a series of unrelated images and 
are cropped into $1024\times 1024$ patches with a stride of 824, obtaining 26155 training samples~\cite{li2023mtb}, 
which follows the DOTA settings~\cite{Ding2021ObjectDI}. 
And all methods are trained for 12 epochs. It is time costly.
In this paper, we reconstruct the OBB benchmark of SAT-MTB and expand the comparison methods to 15.
We feed the original videos to them and randomly select $512\times 512$ patches to train for 7 epochs.
The batch size is 8 and the learning rate is 0.02.
The momentum and weight decay are set to 0.9 and 0.0001, respectively.
In comparison experiments with video detectors, all settings are the same as the HBB benchmark of SAT-MTB~\cite{li2023mtb}.

\subsection{Comparisons with the State-of-the-Art} 
\subsubsection{\textbf{Comparison with OBB detectors}}
\begin{table*}[!thbp] 
  \footnotesize
  \renewcommand\arraystretch{1.5}
  \centering
  \caption{Comparison results with OBB detectors on OBB detection benchmark of SAT-MTB dataset.
  Bolding indicates the highest precision.\label{tab1}}
  \setlength{\tabcolsep}{0.6mm}{
  \begin{tabular}{c|c|c|c|c|c|cccccccccccc|c}
  
  \specialrule{0.1em}{1pt}{1pt}
  \multirow{2}{*}{\textbf{Method}} & {\textbf{Accepted}} & \multirow{2}{*}{\textbf{Backbone}} & {\textbf{Training}} & \multirow{2}{*}{\textbf{Size}} & \multirow{2}{*}{\textbf{Epoch}} & \multicolumn{12}{c|}{$\rm AP$ of Classes} & \multirow{2}{*}{$\rm mAP$} \\
  \cline{7-18}
  {} & {\textbf{By}} & {} & {\textbf{Images}} & {} & {} & {CA} & {CS} & {FA} & {FH} & {NA} & {NV} & {OS} & {RA} & {SB} & {TN} & {WA} & {YC} & {} \\
  \hline
  \multicolumn{19}{c}{\textbf{Official OBB Detection Benchmark Results}} \\
  \hline
  {Faster RCNN-OBB} & {IEEE2017} & \multirow{7}{*}{ResNet50} & \multirow{8}{*}{26155} & \multirow{8}{*}{1024} & \multirow{8}{*}{12} & {20.7} & {40.2} & {86.5} & {0} & {83} & {0} & {55.7} & {83.4} & {10.2} & {0.1} & {85.7} & {31.6} & {41.4} \\
  {RetinaNet-OBB} & {ICCV2017} & {} & {} & {} & {} & {51.9} & {13.2} & {86.8} & {5.8} & {81.3} & {0} & {39.6} & {79.2} & {1.0} & {0} & {84.2} & {22.6} & {38.8} \\
  {RoI Transformer} & {CVPR2018} & {} & {} & {} & {} & {38.2} & {0.6} & {90.6} & {0.8} & {84.6} & {0.1} & {35.5} & {82.7} & {10.4} & {1.1} & {80.3} & {54.1} & {39.9} \\
  {FCOS-OBB} & {ICCV2019} & {} & {} & {} & {} & {42.6} & {0.1} & {90.6} & {0} & {82.9} & {0} & {57} & {85.7} & {3.4} & {0.1} & {80.4} & {40.2} & {40.3} \\
  {Oriented RCNN} & {ICCV2021} & {} & {} & {} & {} & {42.2} & {14.1} & {90} & {0} & {84.4} & {0.3} & {36.4} & {81.1} & {1.2} & {3.6} & {80.4} & {43.8} & {39.8} \\
{R3Det} & {AAAI2021} & {} & {} & {} & {} & {20.7} & {40.2} & {86.5} & {0} & {83} & {0} & {55.7} & {83.4} & {10.2} & {0.1} & {85.7} & {31.6} & {41.4} \\
{S2ANet} & {TGRS2021} & {} & {} & {} & {} & {19.1} & {0.8} & {89.7} & {0} & {75.5} & {0} & {69.9} & {72.7} & {9.4} & {0} & {85.9} & {18.3} & {36.8} \\
\cline{1-3} \cline{7-19}
{ReDet} & {CVPR2021} & {ReResNet50} & {} & {} & {} & {81.1} & {0.2} & {81.8} & {0} & {84.4} & {0.2} & {53.6} & {85.1} & {2.6} & {0.7} & {80.3} & {48.7} & {43.2} \\

  \hline
  \multicolumn{19}{c}{\textbf{OBB Detection Results in Our Settings}} \\
  \hline
  {Faster RCNN-OBB} & {IEEE2017} & \multirow{14}{*}{ResNet50} & \multirow{16}{*}{12573} & \multirow{16}{*}{512} & \multirow{16}{*}{7} & {70.7} & {0} & {89.3} & {0} & {83.6} & {0} & {20.5} & {90} & {11.2} & {1.5} & {77} & {\textbf{50.4}} & {41.2} \\
  {RetinaNet-OBB} & {ICCV2017} & {} & {} & {} & {} & {45.4} & {29.7} & {\textbf{90.2}} & {\textbf{9.6}} & {82.9} & {0} & {4} & {87.9} & {1.6} & {0} & {78.2} & {33.6} & {38.6} \\
  {RoI Transformer} & {CVPR2018} & {} & {} & {} & {} & {64} & {8.1} & {81.7} & {0} & {85.8} & {0} & {33.6} & {89} & {9.7} & {1.5} & {78.1} & {54} & {42.1} \\
  {FCOS-OBB} & {ICCV2019} & {} & {} & {} & {} & {43} & {2.8} & {90.1} & {0} & {81} & {0} & {1.6} & {86.3} & {3.3} & {0} & {76.3} & {30.4} & {34.6} \\
  {RepPoints-OBB} & {ICCV2019} & {} & {} & {} & {} & {51.2} & {0.9} & {87.4} & {0.1} & {84.6} & {0} & {17.6} & {61.4} & {1.5} & {0} & {82.6} & {33.9} & {35.1} \\
  {ATSS-OBB} & {CVPR2020} & {} & {} & {} & {} & {65.5} & {2.1} & {87.9} & {0} & {83} & {0} & {3.6} & {79.5} & {1.4} & {0} & {82.3} & {38.3} & {37} \\
  {Oriented RCNN} & {ICCV2021} & {} & {} & {} & {} & {66.5} & {0.1} & {81.8} & {0} & {85.6} & {0.3} & {34.6} & {82.3} & {1.2} & {\textbf{9.1}} & {78.2} & {45.5} & {40.4} \\
{R3Det} & {AAAI2021} & {} & {} & {} & {} & {74.2} & {\textbf{51.9}} & {88} & {0.1} & {82.5} & {0.1} & {2.2} & {91.2} & {0.7} & {0} & {81.6} & {36.5} & {42.4} \\
{S2ANet} & {TGRS2021} & {} & {} & {} & {} & {68.9} & {21.4} & {86.2} & {0.6} & {81.6} & {0} & {23.7} & {86.4} & {2.7} & {0.8} & {77.2} & {41.5} & {40.9} \\
{Gliding Vertex} & {TPAMI2021} & {} & {} & {} & {} & {70.4} & {6.4} & {81.7} & {0} & {84.8} & {0} & {33.4} & {86.4} & {1.2} & {4.5} & {77.3} & {46.4} & {41.1} \\
{GWD} & {ICML2021} & {} & {} & {} & {} & {59.1} & {4.7} & {86.3} & {0.9} & {85} & {0} & {4} & {86.6} & {1.9} & {0} & {\textbf{83.9}} & {29.8} & {36.8} \\
{KFIoU} & {-} & {} & {} & {} & {} & {69} & {2.5} & {89.2} & {0.6} & {84.2} & {0} & {0} & {\textbf{92.1}} & {1.8} & {\textbf{9.1}} & {82.8} & {33.8} & {38.8} \\
{KLD} & {NIPS2021} & {} & {} & {} & {} & {84.1} & {39.1} & {85.9} & {2.3} & {\textbf{86.5}} & {0} & {0.7} & {90} & {1.6} & {0.2} & {83.1} & {39.4} & {42.7} \\
{SASM} & {AAAI2022} & {} & {} & {} & {} & {72.1} & {4.4} & {84.7} & {0.2} & {82.4} & {\textbf{7}} & {2.8} & {53.4} & {10.6} & {9.4} & {81.9} & {42.7} & {37.6} \\
\cline{1-3} \cline{7-19}
{ReDet(Baseline)} & {CVPR2021} & {ReResNet50} & {} & {} & {} & {85.7} & {0} & {90.1} & {1.3} & {85} & {0.2} & {34.5} & {88.6} & {\textbf{12.3}} & {0} & {82} & {35} & {42.9} \\
\cline{1-3} \cline{7-19}
{TCL(Ours)} & {-} & {ReResNet50} & {} & {} & {} & {\textbf{87.5}} & {0} & {90.1} & {0} & {81.7} & {0} & {\textbf{86.9}} & {88.0} & {11.9} & {\textbf{9.1}} & {71.2} & {45.4} & {\textbf{47.7}} \\

  \specialrule{0.1em}{1pt}{1pt}
  \end{tabular}}
  \end{table*}

  The evaluation is performed on the OBB benchmark of the SAT-MTB dataset, 
  which contains a considerable number of oriented and fine-grained categories, and complexity scenes. 
  The results on the OBB benchmark of SAT-MTB are shown in table~\ref{tab1}. 
  Compared to the official OBB benchmark of the SAT-MTB dataset, different methods achieve comparable and even better results in our settings. 
  But the training processes of our settings are more effective. 
  In the official OBB benchmark of SAT-MTB dataset, the detection accuracies mAP of RoI Transformer, Oriented RCNN, R3Det, S2ANet, ReDet 
  are 39.9$\%$, 39.8$\%$, 41.4$\%$, 36.8$\%$, 43.2$\%$, respectively. 
  While in our settings, their mAPs are 42.1$\%$($\uparrow$2.2$\%$), 40.4$\%$($\uparrow$0.6$\%$), 
  42.4$\%$($\uparrow$1.0$\%$), 40.9$\%$($\uparrow$4.1$\%$), 42.9$\%$($\downarrow$0.3$\%$).

  In our settings, among the 15 compared methods, the DB-FA (detection based on feature alignment) methods 
  (e.g. R3Det, S2ANet, Oriented RCNN)
  exhibit higher mAPs of 42.4$\%$, 40.9$\%$, and 40.4$\%$, respectively.
  The DB-OOR (detection based on oriented object representation) method Gliding Vertex achieves a mAP of 41.1$\%$. 
  For GWD, KLD, and KFIoU, which represent oriented objects using a 2-D Gaussian distribution 
  and employ different loss functions to calculate distances between Gaussian distributions as regression loss, 
  their effectiveness is reduced for small objects (e.g., OS). 
  Notably, KLD outperforms with a higher mAP of 42.7$\%$. 
  This superiority can be attributed to KLD's scale-invariant nature, 
  proving more effective in satellite videos with significant scale variations. 
 
   The DB-RIFL (detection based on rotation-invariant feature learning) method ReDet 
   attains the highest mAP of 42.9$\%$, showcasing its ability to extract completely rotation-invariant features 
   from rotation-equivariant features, making it well-suited for satellite scenes.

  \begin{figure*}[!htbp]
    \centering
    \includegraphics[width=7in]{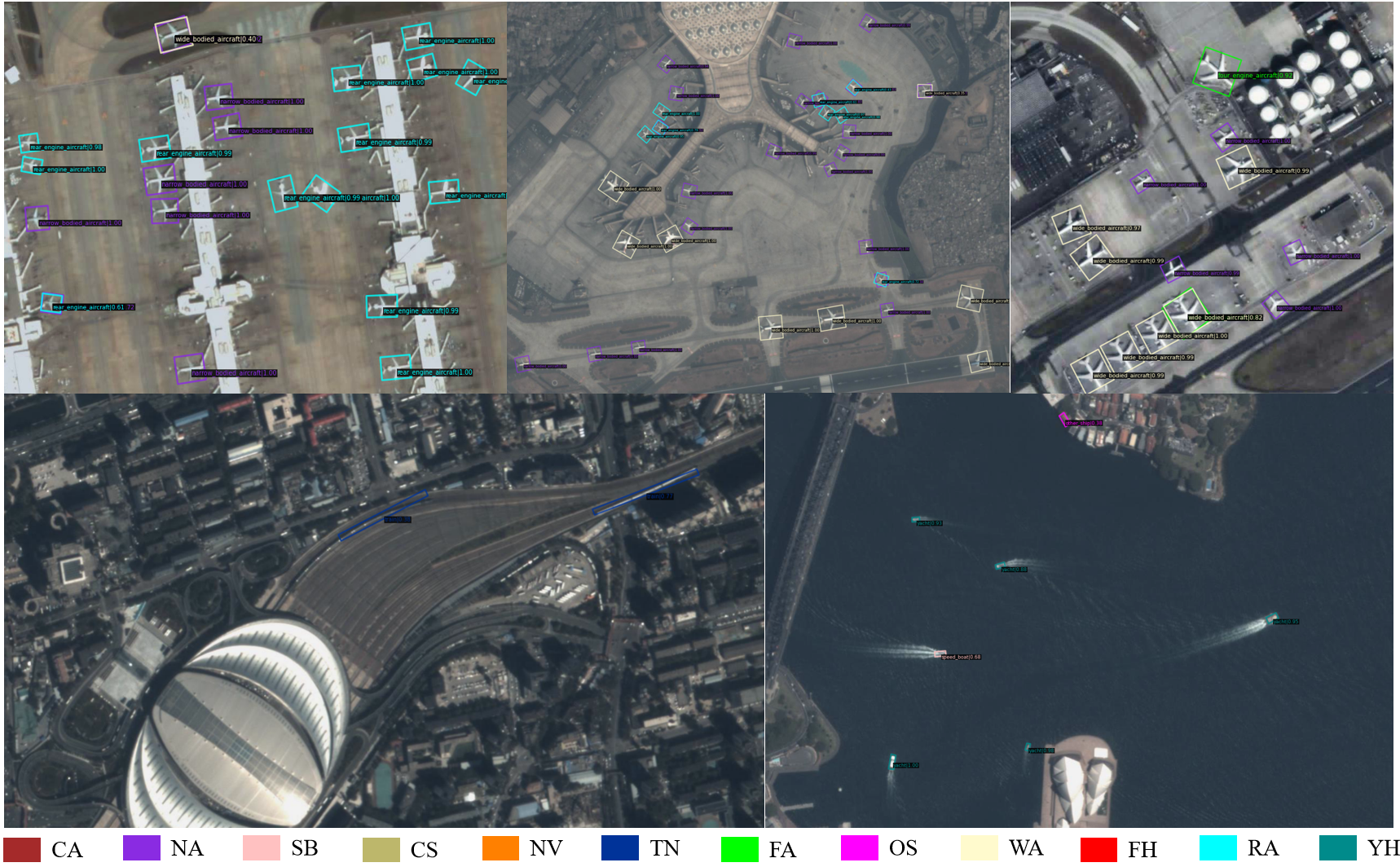}
    \caption{Visualization of satellite video object detection results of OBB benchmark of SAT-MTB dataset. 
    Different colors indicate different fine-grained categories.}
    \label{figrlt}
  \end{figure*}

  In tailored settings, our TCL method emerges as a standout performer, achieving an impressive 47.7$\%$ mAP—outshining all other methods 
  in the comparison. This showcases a substantial 4.8$\%$ improvement over the baseline. 
  Noteworthy is TCL's exceptional capability in detecting categories characterized by small instances, 
  such as OS ($\uparrow$52.4$\%$) and CA ($\uparrow$1.8$\%$), and handling objects 
  with large-scale variations and diverse aspect ratios, as exemplified by TN ($\uparrow$9.1$\%$). 
  The visual results in figure~\ref{figrlt} vividly illustrate our TCL's proficiency in the OBB detection benchmark of SAT-MTB, 
  particularly in capturing small and fine-grained objects within the dynamic context of satellite videos.

\subsubsection{\textbf{Comparison with video detectors}}

  \begin{table*}[!thbp] 
    \footnotesize
    \renewcommand\arraystretch{1.5}
    \centering
    \caption{Comparison results with video detectors on HBB detection benchmark of SAT-MTB dataset.
    Bolding indicates the highest precision.\label{tab2}}
    \setlength{\tabcolsep}{0.6mm}{
    \begin{tabular}{c|c|c|c|c|c|cccccccccccc|c}
    
    \specialrule{0.1em}{1pt}{1pt}
    \multirow{2}{*}{\textbf{Method}} & {\textbf{Accepted}} & {\textbf{Base}} & \multirow{2}{*}{\textbf{Backbone}} & {\textbf{Training}} & \multirow{2}{*}{\textbf{Size}} & \multicolumn{12}{c|}{$\rm AP$ of Classes} & \multirow{2}{*}{$\rm mAP$} \\
    \cline{7-18}
    {} & {\textbf{By}} & {\textbf{Detector}} & {} & {\textbf{Images}} & {} & {FH} & {RA} & {TN} & {NV} & {YC} & {NA} & {CS} & {FA} & {WA} & {SB} & {CA} & {OS} & {} \\
    \hline
    \multicolumn{19}{c}{\textbf{Image-based Detectors}} \\
    \hline
    {Faster RCNN} & {TPAMI2017} & {-} & {ResNet50} & {12573} & {512} & {9.8} & {64.9} & {42.8} & {0} & {32.1} & {\textbf{62.0}} & {0} & {36.6} & {52.2} & {0.7} & {20.8} & {2.2} & {27.0} \\
    \hline
    {YOLOv3 } & {ICCV2018} & {-} & {DarkNet53} & {12573} & {512} & {5.5} & {60.7} & {44.2} & {0} & {29.4} & {48.8} & {0.4} & {\textbf{49.7}} & {50.5} & {8.8} & {30.2} & {\textbf{65.7}} & {32.8} \\
    \hline
    {Grid RCNN} & {CVPR2019} & \multirow{11}{*}{-} & \multirow{11}{*}{ResNet50} & \multirow{11}{*}{12573} & \multirow{11}{*}{512} & {10.4} & {80.5} & {61.5} & {0} & {39.7} & {46.3} & {0} & {28.4} & {56.3} & {1.3} & {34.1} & {10.2} & {30.7} \\
    {TridentNet} & {ICCV2019} & {} & {} & {} & {} & {0} & {66.8} & {52.3} & {0} & {19.1} & {54.4} & {0} & {44.6} & {61.2} & {2.6} & {33.4} & {0} & {27.9} \\
    {FCOS } & {ICCV2019} & {} & {} & {} & {} & {9.6} & {37.6} & {36.8} & {0} & {27.6} & {43.3} & {0} & {25.6} & {52.7} & {5.9} & {12.5} & {11.8} & {21.9} \\
    {Guided Anchoring} & {CVPR2019} & {} & {} & {} & {} & {13.6} & {\textbf{84.2}} & {61.9} & {0} & {42.4} & {50.8} & {0} & {40.6} & {66.3} & {\textbf{14.5}} & {24.2} & {22.8} & {35.1} \\
    {RetinaNet} & {TPAMI2020} & {} & {} & {} & {} & {9.1} & {58.1} & {42.0} & {0} & {24.0} & {49.2} & {0} & {23.3} & {48.8} & {8.5} & {10.5} & {0.3} & {22.8} \\
    {DETR} & {ECCV2020} & {} & {} & {} & {} & {\textbf{43.2}} & {63.9} & {48.9} & {0} & {36.9} & {54.5} & {0.1} & {47.9} & {50.8} & {6.6} & {\textbf{37.2}} & {11.9} & {33.5} \\
    {ATSS} & {CVPR2020} & {} & {} & {} & {} & {31.7} & {54.6} & {37.7} & {0} & {33.0} & {56.0} & {0} & {30.1} & {57.2} & {3.4} & {18.1} & {32.3} & {29.5} \\
    {LibraRCNN} & {IJCV2021} & {} & {} & {} & {} & {32.7} & {47.9} & {52.4} & {0.2} & {33.6} & {43.0} & {0} & {29.9} & {47.9} & {0.1} & {13.8} & {8.9} & {25.8} \\
    {Groie} & {ICPR2021} & {} & {} & {} & {} & {3.3} & {72.8} & {59.2} & {0} & {42.6} & {50.7} & {0.1} & {36.9} & {59} & {8.1} & {24.6} & {13.8} & {30.9} \\
    {Cascade RCNN} & {TPAMI2021} & {} & {} & {} & {} & {24.8} & {61.6} & {55.1} & {0} & {28.3} & {58.7} & {0} & {26.4} & {51.5} & {0.4} & {21.7} & {1.5} & {27.5} \\
    {DINO} & {ICLR2023} & {} & {} & {} & {} & {13.2} & {59.0} & {64.1} & {0} & {33.8} & {59.0} & {0.8} & {28.4} & {51.4} & {7.7} & {2.5} & {39.8} & {30.0} \\

    \hline
    \multicolumn{19}{c}{\textbf{Video Detectors}} \\
    \hline
    {FGFA} & {ICCV2017} & \multirow{4}{*}{FasterRCNN} & \multirow{4}{*}{ResNet50} & \multirow{4}{*}{12573} & \multirow{4}{*}{512} & {12.6} & {69.3} & {34.5} & {\textbf{2.2}} & {5.1} & {42.1} & {1.3} & {42.0} & {68.7} & {0} & {0.7} & {15.9} & {24.5} \\
    {DFF} & {CVPR2017} & {} & {} & {} & {} & {21.2} & {67.1} & {23.8} & {0} & {6.7} & {36.4} & {3.1} & {41.6} & {65.4} & {0} & {0.7} & {28.9} & {24.6} \\
    {SELSA} & {ICCV2019} & {} & {} & {} & {} & {19.9} & {65.1} & {34.3} & {0.2} & {7.3} & {52.3} & {0.6} & {46.5} & {\textbf{71.0}} & {0} & {1.6} & {17.3} & {26.3}  \\
    {TemporalRoI Align} & {AAAI2021} & {} & {} & {} & {} & {6.3} & {44.3} & {22.6} & {0.1} & {3.4} & {35.4} & {\textbf{4.6}} & {33.4} & {55.3} & {0.1} & {0} & {3.1} & {17.4} \\
    \hline
    \multirow{4}{*}{TCL(Ours)} & \multirow{4}{*}{-} & {FasterRCNN} & \multirow{4}{*}{ResNet50} & \multirow{4}{*}{12573} & \multirow{4}{*}{512} & {10.1} & {83.8} & {60.1} & {0} & {43.6} & {48} & {0} & {39.4} & {56} & {\textbf{14.5}} & {27.7} & {27} & {34.2} \\
    {} & {} & {LibraRCNN} & {} & {} & {} & {12.5} & {78.5} & {61.6} & {0} & {\textbf{48.6}} & {55.5} & {0} & {35.3} & {58.1} & {14.1} & {23.3} & {19.3} & {33.9} \\
    {} & {} & {CascadeRCNN} & {} & {} & {} & {10} & {81.1} & {55.9} & {0} & {40.9} & {58.8} & {0} & {30.1} & {62.9} & {3.4} & {35.4} & {47.6} & {35.5} \\
    {} & {} & {Guided Anchoring} & {} & {} & {} & {15.5} & {79.4} & {\textbf{66.2}} & {0} & {44.3} & {55.2} & {0.2} & {43.5} & {67.3} & {9.2} & {26.2} & {29.3} & {\textbf{36.4}} \\
    
    \specialrule{0.1em}{1pt}{1pt}
    \end{tabular}}
    \end{table*}

    In the comparison, video detectors typically utilize Horizontal Bounding Boxes (HBBs) for object localization. 
    To contrast these methods and highlight the temporal context mining capabilities of our TCL method, 
    we assess their performance on the HBB benchmark of the SAT-MTB dataset. 
    Furthermore, we conduct a comprehensive comparison between our TCL and state-of-the-art (SOTA) image-based detectors 
    to validate its effectiveness extensively. It is essential to mention that, in this section, 
    our TCL employs the ResNet50+FPN feature extraction module. 
    Additionally, we replace the Oriented Region Proposal Network (ORPN) with the 
    Region Proposal Network (RPN) in this section.
    As depicted in table~\ref{tab2}, most image-based methods exhibit superior detection accuracy (mAP) compared to video detectors. 
    This discrepancy can be attributed to the fact that existing video detectors are primarily designed for general videos, 
    overlooking the challenges posed by small-scale and large-scale variations in satellite videos. 
    On the contrary, image-based detectors are adept at addressing the complexities of multiscale object detection. 
    Notably, the transformer-based detector DETR achieves the highest detection accuracy, with an mAP of 33.5$\%$, 
    surpassing all other comparison methods.

    Our TCL serves as a video detector framework, and existing image-based detectors can seamlessly integrate into this framework, 
    benefiting from our proposed modules. In the current landscape, video detectors commonly employ Faster RCNN as the base detector. 
    Adhering to this convention, our TCL achieves an mAP of 34.2$\%$, 
    surpassing all comparison video detectors and outperforming the majority of image-based detectors. 
    Notably, compared to the base detector Faster RCNN, our TCL demonstrates a remarkable 7.2$\%$ improvement in mAP, 
    underscoring its superiority in temporal context mining.
    When employing alternative base detectors, such as Libra RCNN, Cascade RCNN, and Guided Anchoring, 
    our TCL achieves mAPs of 33.9$\%$ ($\uparrow$8.1$\%$), 35.5$\%$ ($\uparrow$8.0$\%$), and 36.4$\%$ ($\uparrow$1.3$\%$), respectively. 
    Particularly for fine-grained categories characterized by small instances (e.g., SB, OS), large-scale variation (e.g., CA, RA), 
    and diverse aspect ratios (e.g., TN ($\uparrow$9.1$\%$)), our method exhibits superior performance, 
    highlighting its adaptability and effectiveness.

\subsection{Ablation Studies}
In this section, we perform a series of ablation experiments on the OBB benchmark of the SAT-MTB dataset to 
assess the effectiveness of our proposed TCL method. Our baseline is ReDet, 
with ReResNet serving as the backbone. The training duration for all models is set to 7 epochs.

\begin{table}[!thbp] 
  \footnotesize
  \renewcommand\arraystretch{1.2}
  \centering
  \caption{Comparisons with state-of-the-art object detection and instance segmentation methods on the SAT-MTB dataset.\label{tab3}}
  \setlength{\tabcolsep}{2.5mm}{
  \begin{tabular}{c|ccc|c}
  
  \specialrule{0.1em}{1pt}{1pt}
  {\textbf{Method}} & {\textbf{TFA}} & {\textbf{SE}} & {\textbf{TCC}} & {$\rm \textbf{mAP}$} \\
  \hline
  {ReDet(Baseline)} & {} & {} & {} & {42.9} \\
  \hline
  \multirow{4}{*}{TLC(Ours)} & {\checkmark} & {} & {} & {46.7} \\
  {} & {} & {\checkmark} & {} & {45.5} \\
  {} & {} & {} & {\checkmark} & {46.4} \\
  {} & {\checkmark} & {\checkmark} & {\checkmark} & {47.7} \\
  \specialrule{0.1em}{1pt}{1pt}
  \end{tabular}}
  \end{table}

  \begin{figure*}[!htbp]
    \centering
    \includegraphics[width=7in]{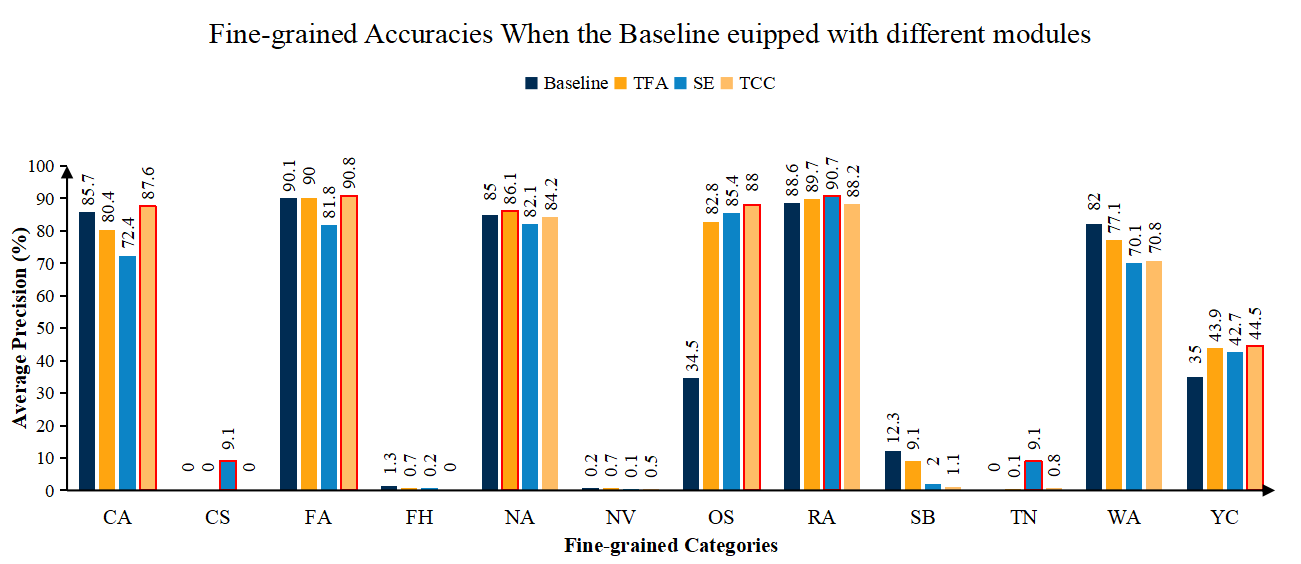}
    \caption{Fine-grained accuracies when the baseline, ReDet, equipped with different modules. 
    The bars with red boundary indicate, comparing with the baseline, the maximum improvement in fine-grained detection accuracies 
    when equipped with our proposed modules.}
    \label{figfap}
  \end{figure*}

  \subsubsection{\textbf{Effectiveness of TFA}}
  \begin{figure}[!htbp]
    \centering
    \includegraphics[width=3in]{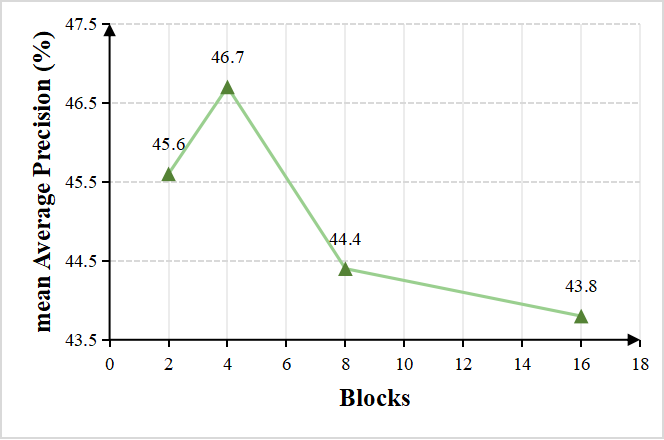}
    \caption{Satellite video object detection accuracies when the fine-grained block number of TFA module changes.}
    \label{figtfa}
  \end{figure}

  To investigate the impact of our Temporal and Fine-grained Feature Aggregation module (TFA),
  we experiment with two settings: with and without TFA. The results presented in table~\ref{tab3} unambiguously demonstrate 
  the necessity of TFA in enhancing the fine-grained detection accuracy of oriented objects.
  TFA yields a notable improvement in mAP by 3.8$\%$ (46.7$\%$ vs 42.9$\%$).
  This substantial increase in accuracy underscores the TFA module's effectiveness in capturing fine-grained features of oriented objects, 
  particularly for categories characterized by small instances (e.g., OS, YC, RA) and large scale variation (e.g., TN, NV).
  For detailed insights into fine-grained detection accuracies, refer to figure~\ref{figfap}.
  
  In TFA, the feature of objects is divided into different blocks for fine-grained feature aggregation in the temporal dimension.
  The number of blocks is a hyperparameter, and for a more nuanced understanding of its impact, we set it to 2, 4, 8, 16, 
  and present the corresponding detection results in figure~\ref{figtfa}.
  As the number of fine-grained blocks increases, the detection accuracy mAP shows an initial rise followed by a decline.
  We attribute this trend to the importance of fine-grained feature aggregation in the temporal dimension.
  However, excessively small local blocks may struggle to represent meaningful fine-grained semantics, leading to noisy feature representation.

  \subsubsection{\textbf{Effectiveness of SE}}
  To assess the importance of the Structure Encoding (SE) module, we conducted a study comparing models with and without the SE module.
  As evident in table~\ref{tab3} and figure~\ref{figfap}, the inclusion of the SE module leads to a significant improvement in detection accuracy, 
  with a gain of 2.6$\%$ (42.9$\%$ vs 45.5$\%$).
  In the context of satellite video oriented and fine-grained object detection, 
  where challenges include small scale, arbitrary angles, and large scale variations, objects often exhibit weak appearance.
  Learning accurate and fine-grained representations from limited and distorted appearance information is challenging.
  The proposed SE module addresses this by providing additional structural information, 
  aiding in distinguishing between different fine-grained categories.
  Notably, the SE module performs particularly well for objects with unique scales, such as CS, OS, RA, and TN.

\subsubsection{\textbf{Effectiveness of TCC}}
\begin{figure*}[!htbp]
  \centering
  \includegraphics[width=6in]{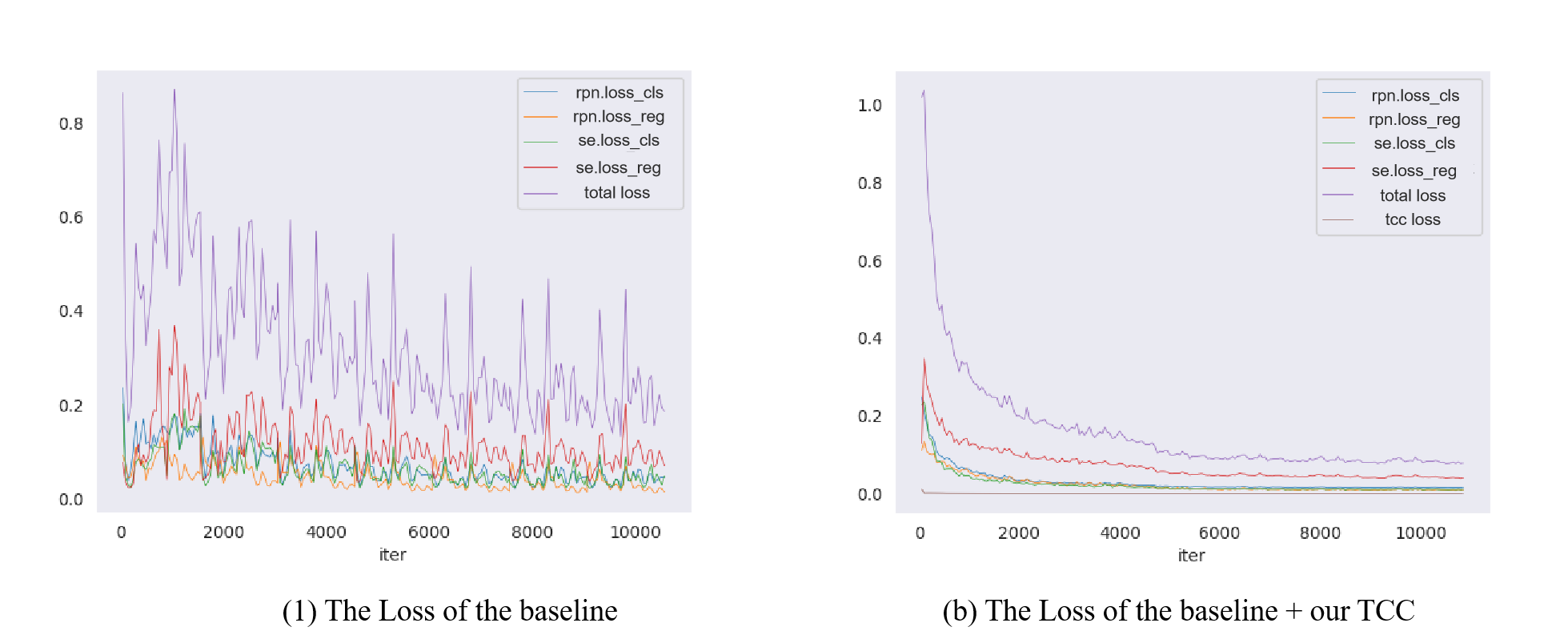}
  \caption{Comparison of the loss changes during training. (a) The loss of the baseline, ReDet. (b) The loss of the baseline + our TCC.}
  \label{figtcc}
\end{figure*}

In this experiment, we investigated the impact of our Temporal Consistency Constraint (TCC) loss.
As outlined in table~\ref{tab3}, the TCC mechanism demonstrated substantial improvements, 
enhancing the detection accuracy mAP by 3.5$\%$ (46.4$\%$ vs 42.9$\%$).
Particularly for objects with challenging feature representations (e.g., CA, OS, TN, YC), our TCL method exhibited superior performance.
The proposed TCC enforces consistency in the fine-grained feature representations of objects across consecutive frames, 
mitigating the impact of inter-frame noise and environmental variations.
This results in a significant boost in detection performance.
Furthermore, a comparison of the losses between the baseline model without TCC and the model with our TCC, 
as shown in figure~\ref{figtcc}, indicates that the model equipped with TCC converges faster and exhibits smoother loss curves.

\begin{figure}[!htbp]
  \centering
  \includegraphics[width=2.5in]{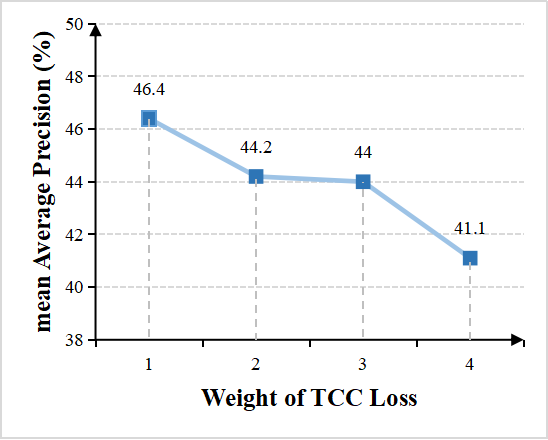}
  \caption{mAPs with the weight changes of TCC loss.}
  \label{figwtcc}
\end{figure}

The weight factor assigned to the TCC loss in the total loss function serves as a hyperparameter.
To explore the impact of this hyperparameter, we present detection results with different weight values (e.g., 1, 2, 3, 4).
As depicted in figure~\ref{figwtcc}, as the weight factor of the TCC loss increases, the mAP decreases.
This suggests that merely increasing the weight of TCC loss in the total loss may not necessarily lead to improved performance.

\section{Conclusion}
Satellite Video Object Detection (SVOD) is crucial for identifying oriented and fine-grained objects in satellite applications. 
Existing SVOD methods often focus on a limited set of coarse-grained moving objects, using horizontal bounding boxes to represent objects. 
This approach struggles to capture complete, accurate, and consistent object information throughout satellite videos, 
posing challenges for practical oriented and fine-grained detection applications.
While some image-based oriented object detectors can be applied directly to satellite videos, 
they treat each frame as an unrelated image, disregarding the rich temporal contexts. 
This fragmentation in object descriptions leads to timing inconsistencies, limiting the effectiveness. 

To address these issues, we propose a novel video object detection framework based on Temporal Consistency Learning (TCL).
Our TCL explores the abundant temporal contexts within satellite videos, incorporating three key modules: 
Temporal and Fine-grained Feature Aggregation (TFA), Structure Encoding (SE), and Temporal Consistency Constraint (TCC). 
To our knowledge, TCL is the first video object detection framework tailored for oriented and fine-grained objects in satellite videos, 
emphasizing temporal context mining.
Our framework detects both moving and static objects across entire satellite video sequences, 
providing a comprehensive representation of their patterns. Unlike image-based detectors that treat each frame independently, 
TCL recognizes oriented and fine-grained objects from the holistic perspective of satellite videos.
Extensive comparison experiments and ablation studies on the largest satellite video benchmark dataset 
demonstrate TCL's superior performance, securing its position atop the leaderboards. 
Furthermore, existing image-based detectors seamlessly adapt to the TCL framework, achieving enhanced detection accuracies.

Looking forward, our future work aims to develop a dedicated object detection network optimized for satellite videos, 
pushing the boundaries of detection performance even further.

\section*{Acknowledgments}
We gratefully acknowledge the support received from the Key Deployment Program of the Chinese Academy of Sciences 
under grant KGFZD-145-23-18. We extend our sincere thanks for the assistance provided through this program.


 

\bibliographystyle{IEEEtran}
\bibliography{TCLbib}

@article{jiao2021new,
  title={New generation deep learning for video object detection: A survey},
  author={Jiao, Licheng and Zhang, Ruohan and Liu, Fang and Yang, Shuyuan and Hou, Biao and Li, Lingling and Tang, Xu},
  journal={IEEE Transactions on Neural Networks and Learning Systems},
  volume={33},
  number={8},
  pages={3195--3215},
  year={2021},
  publisher={IEEE}
}

@article{zhou2023transvod,
  author={Zhou, Qianyu and Li, Xiangtai and He, Lu and Yang, Yibo and Cheng, Guangliang and Tong, Yunhai and Ma, Lizhuang and Tao, Dacheng},
  journal={IEEE Transactions on Pattern Analysis and Machine Intelligence}, 
  title={TransVOD: End-to-End Video Object Detection With Spatial-Temporal Transformers}, 
  year={2023},
  volume={45},
  number={6},
  pages={7853-7869},
  }

@inproceedings{zheng2021survey,
  title={Survey of video object detection algorithms based on deep learning},
  author={Zheng, Lu and Zhou, Tongtong and Jiang, Rongqi and Peng, Yueping},
  booktitle={Proceedings of the 2021 4th International Conference on Algorithms, Computing and Artificial Intelligence},
  pages={1--6},
  year={2021}
}

@article{zhao2019obj,
  author={Zhao, Zhong-Qiu and Zheng, Peng and Xu, Shou-Tao and Wu, Xindong},
  journal={IEEE Transactions on Neural Networks and Learning Systems}, 
  title={Object Detection With Deep Learning: A Review}, 
  year={2019},
  volume={30},
  number={11},
  pages={3212-3232},
  }

@article{zhao2022satsot,
  author={Zhao, Manqi and Li, Shengyang and Xuan, Shiyu and Kou, Longxuan and Gong, Shuai and Zhou, Zhuang},
  journal={IEEE Transactions on Geoscience and Remote Sensing}, 
  title={SatSOT: A Benchmark Dataset for Satellite Video Single Object Tracking}, 
  year={2022},
  volume={60},
  pages={1-11},
  }

@article{li2023mtb,
  author={Li, Shengyang and Zhou, Zhuang and Zhao, Manqi and Yang, Jian and Guo, Weilong and Lv, Yixuan and Kou, Longxuan and Wang, Han and Gu, Yanfeng},
  journal={IEEE Transactions on Geoscience and Remote Sensing}, 
  title={A Multitask Benchmark Dataset for Satellite Video: Object Detection, Tracking, and Segmentation}, 
  year={2023},
  volume={61},
  pages={1-21},
  }

@article{gu2020svsc,
  title={Deep feature extraction and motion representation for satellite video scene classification},
  author={Yanfeng Gu and Huan Liu and Tengfei Wang and Shengyang Li and Guoming Gao},
  journal={Science China Information Sciences},
  year={2020},
  volume={63},
  pages={1-15}
}

@article{li2023satreview,
  author={Li, Shengyang and Sun, Xian and Gu, Yanfeng and Lv, Yixuan and Zhao, Manqi and Zhou, Zhuang and Guo, Weilong and Sun, Yuhan and Wang, Han and Yang, Jian},
  journal={IEEE Journal of Selected Topics in Applied Earth Observations and Remote Sensing}, 
  title={Recent Advances in Intelligent Processing of Satellite Video: Challenges, Methods, and Applications}, 
  year={2023},
  volume={16},
  pages={6776-6798},
  }

@article{FENG2021116,
title = {Cross-frame keypoint-based and spatial motion information-guided networks for moving vehicle detection and tracking in satellite videos},
journal = {ISPRS Journal of Photogrammetry and Remote Sensing},
volume = {177},
pages = {116-130},
year = {2021},
author = {Jie Feng and Dening Zeng and Xiuping Jia and Xiangrong Zhang and Jie Li and Yuping Liang and Licheng Jiao},
}

@article{s23125771,
author = {Li, Ming and Fan, Dazhao and Dong, Yang and Li, Dongzi},
title = {Satellite Video Moving Vehicle Detection and Tracking Based on Spatiotemporal Characteristics},
journal = {Sensors},
volume = {23},
year = {2023},
number = {12},
}

@article{Feng2023SDANetSD,
  title={SDANet: Semantic-Embedded Density Adaptive Network for Moving Vehicle Detection in Satellite Videos},
  author={Jie Feng and Yuping Liang and Xiangrong Zhang and Junpeng Zhang and Licheng Jiao},
  journal={IEEE Transactions on Image Processing},
  year={2023},
  volume={32},
  pages={1788-1801},
}

@article{Xiao2023IncorporatingDB,
  title={Incorporating Deep Background Prior Into Model-Based Method for Unsupervised Moving Vehicle Detection in Satellite Videos},
  author={Chao Xiao and Ting-Yuan Liu and Xinyi Ying and Yingqian Wang and Miao Li and Li Liu and Wei An and Zhijie Chen},
  journal={IEEE Transactions on Geoscience and Remote Sensing},
  year={2023},
  volume={61},
  pages={1-14}
}

@article{Zhang2019ErrorBF,
  title={Error Bounded Foreground and Background Modeling for Moving Object Detection in Satellite Videos},
  author={Junpeng Zhang and Xiuping Jia and Jiankun Hu},
  journal={IEEE Transactions on Geoscience and Remote Sensing},
  year={2019},
  volume={58},
  pages={2659-2669},
}

@article{Zhang2019OnlineSS,
  title={Online Structured Sparsity-Based Moving-Object Detection From Satellite Videos},
  author={Junpeng Zhang and Xiuping Jia and Jiankun Hu and Jocelyn Chanussot},
  journal={IEEE Transactions on Geoscience and Remote Sensing},
  year={2019},
  volume={58},
  pages={6420-6433}
}

@article{Zhang2021MovingVD,
  title={Moving Vehicle Detection for Remote Sensing Video Surveillance With Nonstationary Satellite Platform},
  author={Junpeng Zhang and Xiuping Jia and Jiankun Hu and Kun Tan},
  journal={IEEE Transactions on Pattern Analysis and Machine Intelligence},
  year={2021},
  volume={44},
  pages={5185-5198}
}

@article{Lei2021TinyMV,
  title={Tiny moving vehicle detection in satellite video with constraints of multiple prior information},
  author={Junfeng Lei and Yuxuan Dong and Haigang Sui},
  journal={International Journal of Remote Sensing},
  year={2021},
  volume={42},
  pages={4110 - 4125},
}

@article{Zhang2017SpaceOD,
  title={Space Object Detection in Video Satellite Images Using Motion Information},
  author={Xueyang Zhang and Junhua Xiang and Yulin Zhang},
  journal={International Journal of Aerospace Engineering},
  year={2017},
  volume={2017},
  pages={1-9},
}

@article{Li2019ShipDA,
  title={Ship detection and tracking method for satellite video based on multiscale saliency and surrounding contrast analysis},
  author={Haichao Li and Liang Chen and Feng Li and Meiyu Huang},
  journal={Journal of Applied Remote Sensing},
  year={2019},
  volume={13},
  pages={026511 - 026511},
}

@article{Shi2020DetectingAT,
  title={Detecting and Tracking Moving Airplanes from Space Based on Normalized Frame Difference Labeling and Improved Similarity Measures},
  author={Fan Shi and Fang Qiu and Xiao Li and Ruofei Zhong and Cankun Yang and Yunwei Tang},
  journal={Remote. Sens.},
  year={2020},
  volume={12},
  pages={3589},
}

@article{Shu2021SmallMV,
  title={Small moving vehicle detection via local enhancement fusion for satellite video},
  author={Mengying Shu and Yanfei Zhong and Pengyuan Lv},
  journal={International Journal of Remote Sensing},
  year={2021},
  volume={42},
  pages={7189 - 7214},
}

@article{Chen2020ANA,
  title={A Novel AMS-DAT Algorithm for Moving Vehicle Detection in a Satellite Video},
  author={Xu Chen and Haigang Sui and Jian Fang and Mingting Zhou and Chen Wu},
  journal={IEEE Geoscience and Remote Sensing Letters},
  year={2020},
  volume={19},
  pages={1-5}
}

@inproceedings{Liu2018LowQualityAM,
  title={Low-Quality and Multi-Target Detection in RSIs},
  author={Guiyang Liu and Shengyang Li and Yuyang Shao},
  booktitle={International Conference on Cloud Computing},
  year={2018},
}

@article{Xiao2022DSFNetDA,
  title={DSFNet: Dynamic and Static Fusion Network for Moving Object Detection in Satellite Videos},
  author={Chao Xiao and Qian Yin and Xinyi Ying and Ruojing Li and Shuanglin Wu and Miao Li and Li Liu and Wei An and Zhijie Chen},
  journal={IEEE Geoscience and Remote Sensing Letters},
  year={2022},
  volume={19},
  pages={1-5}
}

@article{Zhou2022FewShotAD,
  title={Few-Shot Aircraft Detection in Satellite Videos Based on Feature Scale Selection Pyramid and Proposal Contrastive Learning},
  author={Zhuang Zhou and Shengyang Li and Weilong Guo and Yanfeng Gu},
  journal={Remote. Sens.},
  year={2022},
  volume={14},
  pages={4581}
}

@article{Pi2022VeryLM,
  title={Very Low-Resolution Moving Vehicle Detection in Satellite Videos},
  author={Zhaoliang Pi and Licheng Jiao and Fang Liu and Xu Liu and Lingling Li and Biao Hou and Shuyuan Yang},
  journal={IEEE Transactions on Geoscience and Remote Sensing},
  year={2022},
  volume={60},
  pages={1-17}
}

@article{Ding2021ObjectDI,
  title={Object Detection in Aerial Images: A Large-Scale Benchmark and Challenges},
  author={Jian Ding and Nan Xue and Guisong Xia and Xiang Bai and Wen Yang and Micheal Ying Yang and Serge J. Belongie and Jiebo Luo and Mihai Datcu and Marcello Pelillo and L. Zhang},
  journal={IEEE Transactions on Pattern Analysis and Machine Intelligence},
  year={2021},
  volume={44},
  pages={7778-7796}
}

@article{Cheng2021AnchorFreeOP,
  title={Anchor-Free Oriented Proposal Generator for Object Detection},
  author={Gong Cheng and Jiabao Wang and Ke Li and Xingxing Xie and Chunbo Lang and Yanqing Yao and Junwei Han},
  journal={IEEE Transactions on Geoscience and Remote Sensing},
  year={2021},
  volume={60},
  pages={1-11}
}

@article{Sun2021FAIR1MAB,
  title={FAIR1M: A Benchmark Dataset for Fine-grained Object Recognition in High-Resolution Remote Sensing Imagery},
  author={Xian Sun and Peijin Wang and Zhiyuan Yan and Feng Xu and Ruiping Wang and Wenhui Diao and Jin Chen and Jihao Li and Yingchao Feng and Tao Xu and Martin Weinmann and Stefan Hinz and Cheng Wang and Kun Fu},
  journal={ArXiv},
  year={2021},
  volume={abs/2103.05569}
}

@article{Wang2023OrientedOD,
  title={Oriented Object Detection in Optical Remote Sensing Images: A Survey},
  author={Kunlin Wang and Zhang Li and Ang Su and Zi Wang},
  journal={ArXiv},
  year={2023},
  volume={abs/2302.10473}
}

@article{LaLonde2017ClusterNetDS,
  title={ClusterNet: Detecting Small Objects in Large Scenes by Exploiting Spatio-Temporal Information},
  author={Rodney LaLonde and Dong Zhang and Mubarak Shah},
  journal={2018 IEEE/CVF Conference on Computer Vision and Pattern Recognition},
  year={2017},
  pages={4003-4012}
}

@article{Li2019WeakMO,
  title={Weak Moving Object Detection In Optical Remote Sensing Video With Motion-Drive Fusion Network},
  author={Yuxuan Li and Licheng Jiao and Xu Tang and Xiangrong Zhang and Wenhua Zhang and Li Gao},
  journal={IGARSS 2019 - 2019 IEEE International Geoscience and Remote Sensing Symposium},
  year={2019},
  pages={5476-5479}
}

@article{Chi2020AerialVM,
  title={Aerial Video Multi-target Detection with Memory Module *},
  author={Haihong Chi and Xiangrui Gao},
  journal={2020 39th Chinese Control Conference (CCC)},
  year={2020},
  pages={7487-7491}
}

@article{Zhang2017ShuffleNetAE,
  title={ShuffleNet: An Extremely Efficient Convolutional Neural Network for Mobile Devices},
  author={Xiangyu Zhang and Xinyu Zhou and Mengxiao Lin and Jian Sun},
  journal={2018 IEEE/CVF Conference on Computer Vision and Pattern Recognition},
  year={2017},
  pages={6848-6856}
}

@article{Xu2019GlidingVO,
  title={Gliding Vertex on the Horizontal Bounding Box for Multi-Oriented Object Detection},
  author={Yongchao Xu and Mingtao Fu and Qimeng Wang and Yukang Wang and Kai Chen and Guisong Xia and Xiang Bai},
  journal={IEEE Transactions on Pattern Analysis and Machine Intelligence},
  year={2019},
  volume={43},
  pages={1452-1459}
}

@article{Wang2020LearningCP,
  title={Learning Center Probability Map for Detecting Objects in Aerial Images},
  author={Jinwang Wang and Wen Yang and Hengchao Li and Haijian Zhang and Guisong Xia},
  journal={IEEE Transactions on Geoscience and Remote Sensing},
  year={2020},
  volume={59},
  pages={4307-4323}
}

@inproceedings{yang2021rethinking,
  title={Rethinking rotated object detection with gaussian wasserstein distance loss},
  author={Yang, Xue and Yan, Junchi and Ming, Qi and Wang, Wentao and Zhang, Xiaopeng and Tian, Qi},
  booktitle={International conference on machine learning},
  pages={11830--11841},
  year={2021},
  organization={PMLR}
}

@article{yang2021learning,
  title={Learning high-precision bounding box for rotated object detection via kullback-leibler divergence},
  author={Yang, Xue and Yang, Xiaojiang and Yang, Jirui and Ming, Qi and Wang, Wentao and Tian, Qi and Yan, Junchi},
  journal={Advances in Neural Information Processing Systems},
  volume={34},
  pages={18381--18394},
  year={2021}
}

@inproceedings{yang2022kfiou,
  title={The KFIoU Loss for Rotated Object Detection},
  author={Yang, Xue and Zhou, Yue and Zhang, Gefan and Yang, Jirui and Wang, Wentao and Yan, Junchi and ZHANG, XIAOPENG and Tian, Qi},
  booktitle={The Eleventh International Conference on Learning Representations},
  year={2022}
}

@article{Li2021OrientedRF,
  title={Oriented RepPoints for Aerial Object Detection},
  author={Wentong Li and Jianke Zhu},
  journal={2022 IEEE/CVF Conference on Computer Vision and Pattern Recognition (CVPR)},
  year={2021},
  pages={1819-1828}
}

@article{Guo2021BeyondBC,
  title={Beyond Bounding-Box: Convex-hull Feature Adaptation for Oriented and Densely Packed Object Detection},
  author={Zonghao Guo and Chang Liu and Xiaosong Zhang and Jianbin Jiao and Xiangyang Ji and Qixiang Ye},
  journal={2021 IEEE/CVF Conference on Computer Vision and Pattern Recognition (CVPR)},
  year={2021},
  pages={8788-8797}
}

@article{Hou2022GRepGR,
  title={G-Rep: Gaussian Representation for Arbitrary-Oriented Object Detection},
  author={Liping Hou and Ke Lu and Xue Yang and Yuqiu Li and Jian Xue},
  journal={Remote. Sens.},
  year={2022},
  volume={15},
  pages={757}
}

@inproceedings{Azimi2018TowardsMO,
  title={Towards Multi-class Object Detection in Unconstrained Remote Sensing Imagery},
  author={Seyed Majid Azimi and Eleonora Vig and Reza Bahmanyar and Marco K{\"o}rner and Peter Reinartz},
  booktitle={Asian Conference on Computer Vision},
  year={2018}
}

@article{Ma2017ArbitraryOrientedST,
  title={Arbitrary-Oriented Scene Text Detection via Rotation Proposals},
  author={Jianqi Ma and Weiyuan Shao and Hao Ye and Li Wang and Hong Wang and Yingbin Zheng and X. Xue},
  journal={IEEE Transactions on Multimedia},
  year={2017},
  volume={20},
  pages={3111-3122}
}

@article{Zhang2018TowardAS,
  title={Toward Arbitrary-Oriented Ship Detection With Rotated Region Proposal and Discrimination Networks},
  author={Zenghui Zhang and Weiwei Guo and Shengnan Zhu and Wenxian Yu},
  journal={IEEE Geoscience and Remote Sensing Letters},
  year={2018},
  volume={15},
  pages={1745-1749}
}

@article{Ding2019LearningRT,
  title={Learning RoI Transformer for Oriented Object Detection in Aerial Images},
  author={Jian Ding and Nan Xue and Yang Long and Guisong Xia and Qikai Lu},
  journal={2019 IEEE/CVF Conference on Computer Vision and Pattern Recognition (CVPR)},
  year={2019},
  pages={2844-2853}
}

@article{Xie2021OrientedRF,
  title={Oriented R-CNN for Object Detection},
  author={Xingxing Xie and Gong Cheng and Jiabao Wang and Xiwen Yao and Junwei Han},
  journal={2021 IEEE/CVF International Conference on Computer Vision (ICCV)},
  year={2021},
  pages={3500-3509}
}

@article{Yang2019R3DetRS,
  title={R3Det: Refined Single-Stage Detector with Feature Refinement for Rotating Object},
  author={Xue Yang and Qingqing Liu and Junchi Yan and Ang Li},
  journal={ArXiv},
  year={2019},
  volume={abs/1908.05612}
}

@article{Han2020AlignDF,
  title={Align Deep Features for Oriented Object Detection},
  author={Jiaming Han and Jian Ding and Jie Li and Guisong Xia},
  journal={IEEE Transactions on Geoscience and Remote Sensing},
  year={2020},
  volume={60},
  pages={1-11}
}

@article{Pan2020DynamicRN,
  title={Dynamic Refinement Network for Oriented and Densely Packed Object Detection},
  author={Xingjia Pan and Yuqiang Ren and Kekai Sheng and Weiming Dong and Haolei Yuan and Xiao-Wei Guo and Chongyang Ma and Changsheng Xu},
  journal={2020 IEEE/CVF Conference on Computer Vision and Pattern Recognition (CVPR)},
  year={2020},
  pages={11204-11213}
}

@article{Han2021ReDetAR,
  title={ReDet: A Rotation-equivariant Detector for Aerial Object Detection},
  author={Jiaming Han and Jian Ding and Nan Xue and Guisong Xia},
  journal={2021 IEEE/CVF Conference on Computer Vision and Pattern Recognition (CVPR)},
  year={2021},
  pages={2785-2794}
}

@inproceedings{yang2020arbitrary,
  title={Arbitrary-oriented object detection with circular smooth label},
  author={Yang, Xue and Yan, Junchi},
  booktitle={Computer Vision--ECCV 2020: 16th European Conference, Glasgow, UK, August 23--28, 2020, Proceedings, Part VIII 16},
  pages={677--694},
  year={2020},
  organization={Springer}
}

@inproceedings{Hou2022ShapeAdaptiveSA,
  title={Shape-Adaptive Selection and Measurement for Oriented Object Detection},
  author={Liping Hou and Ke Lu and Jian Xue and Yuqiu Li},
  booktitle={AAAI Conference on Artificial Intelligence},
  year={2022}
}

@article{Han2016SeqNMSFV,
  title={Seq-NMS for Video Object Detection},
  author={Wei Han and Pooya Khorrami and Tom Le Paine and Prajit Ramachandran and Mohammad Babaeizadeh and Humphrey Shi and Jianan Li and Shuicheng Yan and Thomas S. Huang},
  journal={ArXiv},
  year={2016},
  volume={abs/1602.08465}
}

@article{Kang2016TCNNTW,
  title={T-CNN: Tubelets With Convolutional Neural Networks for Object Detection From Videos},
  author={Kai Kang and Hongsheng Li and Junjie Yan and Xingyu Zeng and Binh Yang and Tong Xiao and Cong Zhang and Zhe Wang and Ruohui Wang and Xiaogang Wang and Wanli Ouyang},
  journal={IEEE Transactions on Circuits and Systems for Video Technology},
  year={2016},
  volume={28},
  pages={2896-2907}
}

@inproceedings{Belhassen2019ImprovingVO,
  title={Improving Video Object Detection by Seq-Bbox Matching},
  author={Hatem Belhassen and Heng Zhang and Virginie Fresse and El-Bay Bourennane},
  booktitle={VISIGRAPP},
  year={2019}
}

@article{Sabater2020RobustAE,
  title={Robust and efficient post-processing for video object detection},
  author={Alberto Sabater and Luis Montesano and Ana Cristina Murillo},
  journal={2020 IEEE/RSJ International Conference on Intelligent Robots and Systems (IROS)},
  year={2020},
  pages={10536-10542}
}

@inproceedings{Yao2020VideoOD,
  title={Video Object Detection via Object-Level Temporal Aggregation},
  author={Chun-Han Yao and Chen Fang and Xiaohui Shen and Yangyue Wan and Ming-Hsuan Yang},
  booktitle={European Conference on Computer Vision},
  year={2020}
}

@inproceedings{Jiang2019LearningWT,
  title={Learning Where to Focus for Efficient Video Object Detection},
  author={Zhengkai Jiang and Yu Liu and Ceyuan Yang and Jihao Liu and Peng Gao and Qian Zhang and Shiming Xiang and Chunhong Pan},
  booktitle={European Conference on Computer Vision},
  year={2019}
}

@inproceedings{Han2020MiningIP,
  title={Mining Inter-Video Proposal Relations for Video Object Detection},
  author={Mingfei Han and Yali Wang and Xiaojun Chang and Y. Qiao},
  booktitle={European Conference on Computer Vision},
  year={2020}
}

@article{Han2020ExploitingBF,
  title={Exploiting Better Feature Aggregation for Video Object Detection},
  author={Liang Han and Pichao Wang and Zhaozheng Yin and F. Wang and Hao Li},
  journal={Proceedings of the 28th ACM International Conference on Multimedia},
  year={2020}
}

@article{Lin2020DualSF,
  title={Dual Semantic Fusion Network for Video Object Detection},
  author={Lijian Lin and Haosheng Chen and Honglun Zhang and Jun Liang and Yu Li and Ying Shan and Hanzi Wang},
  journal={Proceedings of the 28th ACM International Conference on Multimedia},
  year={2020}
}

@inproceedings{He2020TemporalCE,
  title={Temporal Context Enhanced Feature Aggregation for Video Object Detection},
  author={Fei He and Naiyu Gao and Qiaozhe Li and Senyao Du and Xin Zhao and Kaiqi Huang},
  booktitle={AAAI Conference on Artificial Intelligence},
  year={2020}
}

@article{Chen2018OptimizingVO,
  title={Optimizing Video Object Detection via a Scale-Time Lattice},
  author={Kai Chen and Jiaqi Wang and Shuo Yang and Xingcheng Zhang and Yuanjun Xiong and Chen Change Loy and Dahua Lin},
  journal={2018 IEEE/CVF Conference on Computer Vision and Pattern Recognition},
  year={2018},
  pages={7814-7823}
}

@inproceedings{Wang2018FullyMN,
  title={Fully Motion-Aware Network for Video Object Detection},
  author={Shiyao Wang and Yucong Zhou and Junjie Yan and Zhidong Deng},
  booktitle={European Conference on Computer Vision},
  year={2018}
}

@article{Zhu2017TowardsHP,
  title={Towards High Performance Video Object Detection},
  author={Xizhou Zhu and Jifeng Dai and Lu Yuan and Yichen Wei},
  journal={2018 IEEE/CVF Conference on Computer Vision and Pattern Recognition},
  year={2017},
  pages={7210-7218}
}

@article{jin2022feature,
  title={Feature flow: In-network feature flow estimation for video object detection},
  author={Jin, Ruibing and Lin, Guosheng and Wen, Changyun and Wang, Jianliang and Liu, Fayao},
  journal={Pattern Recognition},
  volume={122},
  pages={108323},
  year={2022}
}

@article{Deng2019ObjectGE,
  title={Object Guided External Memory Network for Video Object Detection},
  author={Hanming Deng and Yang Hua and Tao Song and Zongpu Zhang and Zhengui Xue and Ruhui Ma and Neil Martin Robertson and Haibing Guan},
  journal={2019 IEEE/CVF International Conference on Computer Vision (ICCV)},
  year={2019},
  pages={6677-6686}
}

@article{Guo2019ProgressiveSL,
  title={Progressive Sparse Local Attention for Video Object Detection},
  author={Chaoxu Guo and Bin Fan and Jie Gu and Qian Zhang and Shiming Xiang and V{\'e}ronique Prinet and Chunhong Pan},
  journal={2019 IEEE/CVF International Conference on Computer Vision (ICCV)},
  year={2019},
  pages={3908-3917}
}

@inproceedings{Jiang2019VideoOD,
  title={Video Object Detection with Locally-Weighted Deformable Neighbors},
  author={Zhengkai Jiang and Peng Gao and Chaoxu Guo and Qian Zhang and Shiming Xiang and Chunhong Pan},
  booktitle={AAAI Conference on Artificial Intelligence},
  year={2019}
}

@article{Chen2020MemoryEG,
  title={Memory Enhanced Global-Local Aggregation for Video Object Detection},
  author={Yihong Chen and Yue Cao and Han Hu and Liwei Wang},
  journal={2020 IEEE/CVF Conference on Computer Vision and Pattern Recognition (CVPR)},
  year={2020},
  pages={10334-10343}
}

@article{Deng2019RelationDN,
  title={Relation Distillation Networks for Video Object Detection},
  author={Jiajun Deng and Yingwei Pan and Ting Yao and Wen-gang Zhou and Houqiang Li and Tao Mei},
  journal={2019 IEEE/CVF International Conference on Computer Vision (ICCV)},
  year={2019},
  pages={7022-7031}
}

@article{Wu2019SequenceLS,
  title={Sequence Level Semantics Aggregation for Video Object Detection},
  author={Haiping Wu and Yuntao Chen and Naiyan Wang and Zhaoxiang Zhang},
  journal={2019 IEEE/CVF International Conference on Computer Vision (ICCV)},
  year={2019},
  pages={9216-9224}
}

@inproceedings{Vaswani2017AttentionIA,
  title={Attention is All you Need},
  author={Ashish Vaswani and Noam M. Shazeer and Niki Parmar and Jakob Uszkoreit and Llion Jones and Aidan N. Gomez and Lukasz Kaiser and Illia Polosukhin},
  booktitle={Neural Information Processing Systems},
  year={2017}
}

@article{Wang2017NonlocalNN,
  title={Non-local Neural Networks},
  author={X. Wang and Ross B. Girshick and Abhinav Kumar Gupta and Kaiming He},
  journal={2018 IEEE/CVF Conference on Computer Vision and Pattern Recognition},
  year={2017},
  pages={7794-7803}
}

@article{Li2021ImprovingVI,
  title={Improving Video Instance Segmentation via Temporal Pyramid Routing},
  author={Xiangtai Li and Hao He and Henghui Ding and Kuiyuan Yang and Guangliang Cheng and Jianping Shi and Yunhai Tong},
  journal={IEEE Transactions on Pattern Analysis and Machine Intelligence},
  year={2021},
  volume={45},
  pages={6594-6601}
}

@article{He2022TransVODEV,
  title={TransVOD: End-to-End Video Object Detection With Spatial-Temporal Transformers},
  author={Lu He and Qianyu Zhou and Xiangtai Li and Li Niu and Guangliang Cheng and Xiao Li and Wenxuan Liu and Yunhai Tong and Lizhuang Ma and Liqing Zhang},
  journal={IEEE Transactions on Pattern Analysis and Machine Intelligence},
  year={2022},
  volume={45},
  pages={7853-7869}
}

@article{Ren2015FasterRT,
  title={Faster R-CNN: Towards Real-Time Object Detection with Region Proposal Networks},
  author={Shaoqing Ren and Kaiming He and Ross B. Girshick and Jian Sun},
  journal={IEEE Transactions on Pattern Analysis and Machine Intelligence},
  year={2017},
  volume={39},
  pages={1137-1149}
}

@article{Lin2017FocalLF,
  title={Focal Loss for Dense Object Detection},
  author={Tsung-Yi Lin and Priya Goyal and Ross B. Girshick and Kaiming He and Piotr Doll{\'a}r},
  journal={2017 IEEE International Conference on Computer Vision (ICCV)},
  year={2017},
  pages={2999-3007}
}

@article{Tian2019FCOSFC,
  title={FCOS: Fully Convolutional One-Stage Object Detection},
  author={Zhi Tian and Chunhua Shen and Hao Chen and Tong He},
  journal={2019 IEEE/CVF International Conference on Computer Vision (ICCV)},
  year={2019},
  pages={9626-9635}
}

@article{Yang2019RepPointsPS,
  title={RepPoints: Point Set Representation for Object Detection},
  author={Ze Yang and Shaohui Liu and Han Hu and Liwei Wang and Stephen Lin},
  journal={2019 IEEE/CVF International Conference on Computer Vision (ICCV)},
  year={2019},
  pages={9656-9665}
}

@article{Zhang2019BridgingTG,
  title={Bridging the Gap Between Anchor-Based and Anchor-Free Detection via Adaptive Training Sample Selection},
  author={Shifeng Zhang and Cheng Chi and Yongqiang Yao and Zhen Lei and Stan Z. Li},
  journal={2020 IEEE/CVF Conference on Computer Vision and Pattern Recognition (CVPR)},
  year={2019},
  pages={9756-9765}
}

@inproceedings{zhu2017fgfa,
  title={Flow-guided feature aggregation for video object detection},
  author={Zhu, Xizhou and Wang, Yujie and Dai, Jifeng and Yuan, Lu and Wei, Yichen},
  booktitle={Proceedings of the IEEE international conference on computer vision},
  pages={408--417},
  year={2017}
}

@inproceedings{zhu2017dff,
  title={Deep feature flow for video recognition},
  author={Zhu, Xizhou and Xiong, Yuwen and Dai, Jifeng and Yuan, Lu and Wei, Yichen},
  booktitle={Proceedings of the IEEE conference on computer vision and pattern recognition},
  pages={2349--2358},
  year={2017}
}

@inproceedings{gong2021selsa,
  title={Temporal ROI align for video object recognition},
  author={Gong, Tao and Chen, Kai and Wang, Xinjiang and Chu, Qi and Zhu, Feng and Lin, Dahua and Yu, Nenghai and Feng, Huamin},
  booktitle={Proceedings of the AAAI Conference on Artificial Intelligence},
  volume={35},
  number={2},
  pages={1442--1450},
  year={2021}
}

@article{Redmon2018YOLOv3AI,
  title={YOLOv3: An Incremental Improvement},
  author={Joseph Redmon and Ali Farhadi},
  journal={ArXiv},
  year={2018},
  volume={abs/1804.02767}
}

@inproceedings{Lu2018GridR,
  title={Grid r-cnn},
  author={Lu, Xin and Li, Buyu and Yue, Yuxin and Li, Quanquan and Yan, Junjie},
  booktitle={Proceedings of the IEEE/CVF Conference on Computer Vision and Pattern Recognition},
  pages={7363--7372},
  year={2019}
}

@article{Li2019ScaleAwareTN,
  title={Scale-Aware Trident Networks for Object Detection},
  author={Yanghao Li and Yuntao Chen and Naiyan Wang and Zhaoxiang Zhang},
  journal={2019 IEEE/CVF International Conference on Computer Vision (ICCV)},
  year={2019},
  pages={6053-6062}
}

@article{Wang2019RegionPB,
  title={Region Proposal by Guided Anchoring},
  author={Jiaqi Wang and Kai Chen and Shuo Yang and Chen Change Loy and Dahua Lin},
  journal={2019 IEEE/CVF Conference on Computer Vision and Pattern Recognition (CVPR)},
  year={2019},
  pages={2960-2969}
}

@article{Carion2020EndtoEndOD,
  title={End-to-End Object Detection with Transformers},
  author={Nicolas Carion and Francisco Massa and Gabriel Synnaeve and Nicolas Usunier and Alexander Kirillov and Sergey Zagoruyko},
  journal={ArXiv},
  year={2020},
  volume={abs/2005.12872}
}

@article{Pang2019LibraRT,
  title={Libra R-CNN: Towards Balanced Learning for Object Detection},
  author={Jiangmiao Pang and Kai Chen and Jianping Shi and Huajun Feng and Wanli Ouyang and Dahua Lin},
  journal={2019 IEEE/CVF Conference on Computer Vision and Pattern Recognition (CVPR)},
  year={2019},
  pages={821-830}
}

@article{Rossi2020ANR,
  title={A Novel Region of Interest Extraction Layer for Instance Segmentation},
  author={L. Rossi and Akbar Karimi and Andrea Prati},
  journal={2020 25th International Conference on Pattern Recognition (ICPR)},
  year={2020},
  pages={2203-2209}
}

@article{Cai2019CascadeRH,
  title={Cascade R-CNN: High Quality Object Detection and Instance Segmentation},
  author={Zhaowei Cai and Nuno Vasconcelos},
  journal={IEEE Transactions on Pattern Analysis and Machine Intelligence},
  year={2021},
  volume={43},
  number={5},
  pages={1483-1498}
}

@article{Zhang2022DINODW,
  title={DINO: DETR with Improved DeNoising Anchor Boxes for End-to-End Object Detection},
  author={Hao Zhang and Feng Li and Siyi Liu and Lei Zhang and Hang Su and Jun-Juan Zhu and Lionel Ming-shuan Ni and Heung-yeung Shum},
  journal={ArXiv},
  year={2022},
  volume={abs/2203.03605},
}

@article{Dai2022AO2DETRAO,
  title={AO2-DETR: Arbitrary-Oriented Object Detection Transformer},
  author={Linhui Dai and Hong Liu and Haoling Tang and Zhiwei Wu and Pinhao Song},
  journal={IEEE Transactions on Circuits and Systems for Video Technology},
  year={2022},
  volume={33},
  pages={2342-2356}
}


 
\vspace{11pt}

\begin{IEEEbiography}[{\includegraphics[width=1in,height=1.25in,clip,keepaspectratio]{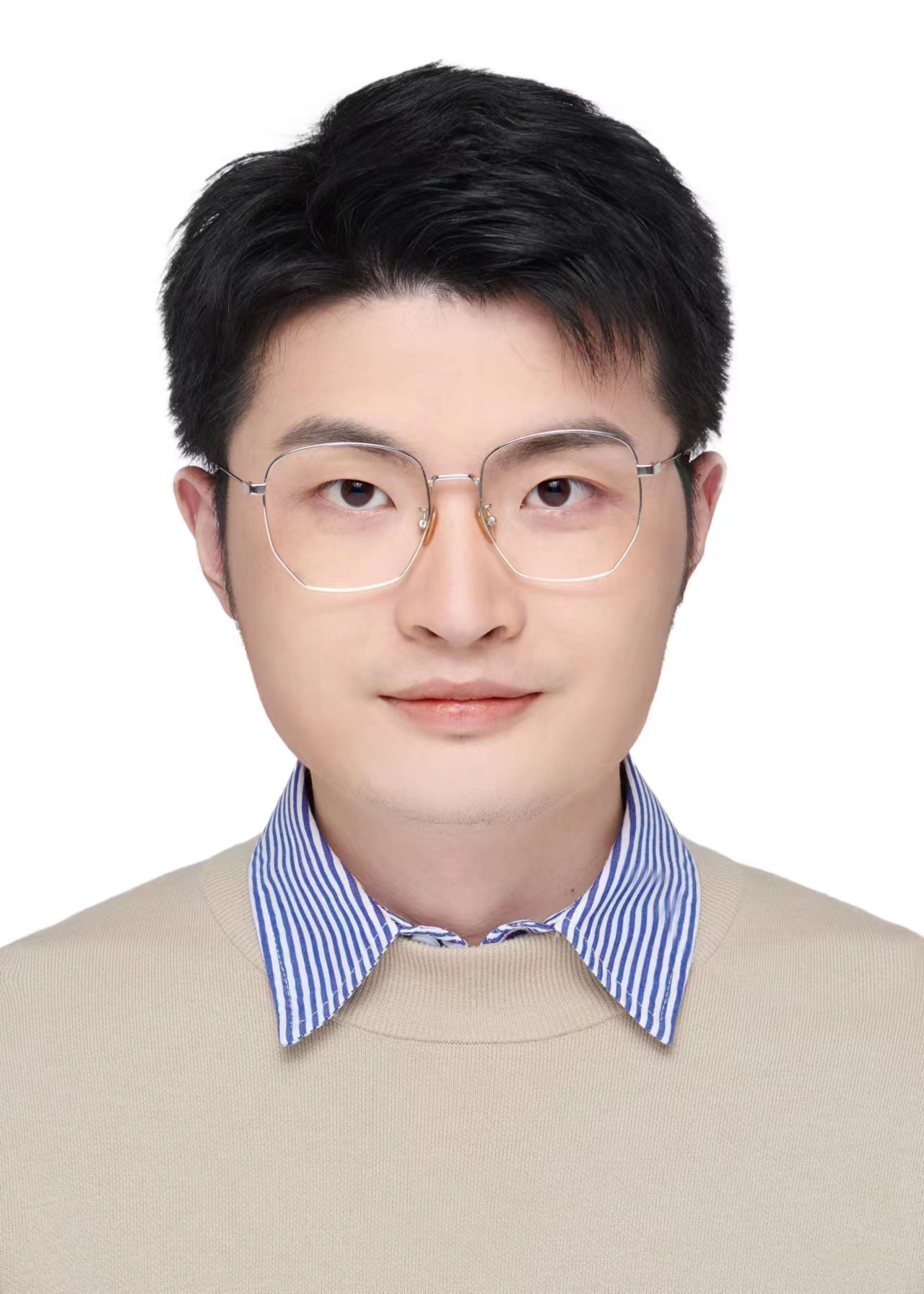}}]{Weilong Guo}
  earned his B.Eng. degree in software engineering from Jilin University, Jilin, China, in 2018, and subsequently, 
  his M.Eng. degree in computer applied technology from the School of Artificial Intelligence, University of Chinese Academy of Sciences, 
  Beijing, China, in 2022.

  Currently serving as an Engineer at the Technology and Engineering Center for Space Utilization, Chinese Academy of Sciences, Beijing, China, 
  he is deeply engaged in research focused on the intelligent analysis and understanding of images and videos.
\end{IEEEbiography}
\begin{IEEEbiography}[{\includegraphics[width=1in,height=1.25in,clip,keepaspectratio]{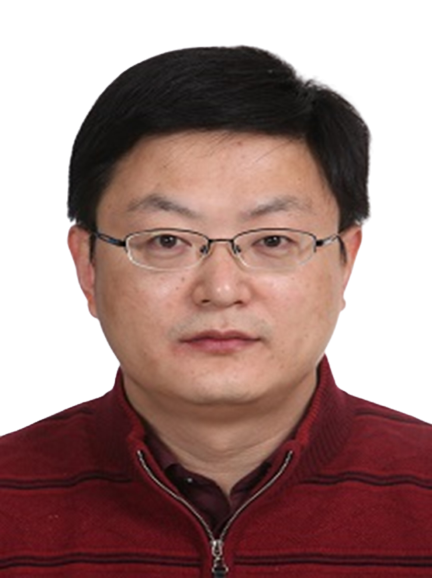}}]{Shengyang Li}
 received the Ph.D. degree from the Institute of Remote Sensing Applications, Chinese Academy of Sciences, Beijing, China, in 2006. 
 
 He is currently a Professor with the Technology and Engineering Center for Space Utilization, Chinese Academy of Sciences. 
 His research activities are machine learning in remote sensing image interpretation, deep learning in satellite 
 videos processing and analysis, intelligent image processing, analysis and understanding for space utilization , 
 and space scientific big data modeling and analysis.
\end{IEEEbiography}
\begin{IEEEbiography}[{\includegraphics[width=1in,height=1.25in,clip,keepaspectratio]{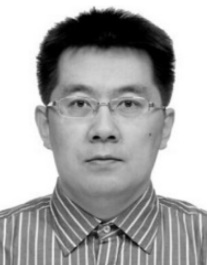}}]{Yanfeng Gu}
  (Senior Member, IEEE) received the Ph.D. degree in information and communication engineering from Harbin Institute of Technology (HIT), 
  Harbin, China, in 2005. 
  
  He joined the School of Electronics and Information Engineering, HIT, as a Lecturer. He was appointed as an Associate Professor at 
  HIT in 2006; meanwhile, he was enrolled in the first Outstanding Younger Teacher Training Program of HIT. From 2011 to 2012, 
  he was a Visiting Scholar with the Department of Electrical Engineering and Computer Science, University of California at Berkeley, 
  Berkeley, CA, USA. He is currently a Professor with the Department of Information Engineering, HIT. He has published more than 100 
  peer-reviewed papers and four book chapters. He is the inventor or a co-inventor of 20 patents. His research interests include space-intelligent 
  remote sensing and information processing, multimodal hyperspectral remote sensing, and space borne time-series image processing.
\end{IEEEbiography}

\vspace{11pt}

\vfill

\end{document}